\documentclass{article}

\PassOptionsToPackage{numbers,sort&compress}{natbib}

\usepackage[preprint]{neurips_2023}

\usepackage{microtype}
\usepackage{graphicx}
\usepackage{booktabs} %
\usepackage{tabularx}
\usepackage{wrapfig} %
\usepackage{multirow}
\usepackage[font=footnotesize]{caption}
\usepackage[font=footnotesize]{subcaption}

\usepackage{hyperref}
\usepackage{url}            %

\usepackage[linesnumbered,ruled]{algorithm2e}
\usepackage{floatrow}
\newfloatcommand{capbtabbox}{table}[][\FBwidth]

\usepackage{amsmath}
\usepackage{amssymb}
\usepackage{mathtools}
\usepackage{amsthm}
\usepackage{delimset}

\usepackage{amsfonts, bbm}       %
\usepackage{nicefrac}       %
\usepackage{xcolor}         %
\usepackage{tikz}
\usetikzlibrary{positioning,arrows.meta}

\definecolor{TabBlue}{RGB}{31,119,180}
\definecolor{TabRed}{RGB}{214,39,40}
\definecolor{TabGreen}{HTML}{2ca02c}
\definecolor{light-gray}{gray}{0.95}
\definecolor{light-blue}{HTML}{e9f4fb}
\definecolor{light-red}{HTML}{fbe9e9}

\usepackage{pifont}%

\usepackage[capitalize,noabbrev]{cleveref}

\theoremstyle{plain}
\newtheorem{theorem}{Theorem}[section]

\theoremstyle{definition}
\newtheorem{definition}[theorem]{Definition}

\theoremstyle{remark}

\usepackage[textsize=tiny]{todonotes} %

\usepackage{comment}

\def\showComments{} %

\ifdefined\showComments
    \newcommand{\guy}[1]{ \textcolor{blue}{\footnotesize \{Guy: #1\}}}
    \newcommand{\alizee}[1]{\textcolor{red}{\footnotesize \{Alizee: #1\}}}
\else
    \newcommand{\guy}[1]{}
    \newcommand{\alizee}[1]{}
\fi

\title{Delphic Offline Reinforcement Learning \\ under
Nonidentifiable Hidden Confounding}

\newcommand{\dataset}{\mathcal{D}}
\newcommand{\behav}{\pi_{b}} %

\newcommand{\Sspace}{\mathcal{S}}
\newcommand{\Zspace}{\mathcal{Z}}
\newcommand{\Aspace}{\mathcal{A}}
\newcommand{\M}{\mathcal{M}}
\newcommand{\W}{\mathcal{W}}
\newcommand{\Expt}{\mathbb{E}}

\newcommand*\wc{{}\cdot{}}
\newcommand{\Var}{\text{Var}}

\DeclareMathOperator*{\argmax}{arg\,max}
\DeclareMathOperator*{\argmin}{arg\,min}

\usepackage{todonotes}

\author{%
  Alizée Pace\,$^{1,2,3}$ \qquad \qquad Hugo Yèche\,$^{2}$ \\
  \\ 
  \\
  \textbf{Bernhard Schölkopf}\,$^{3}$ \qquad \quad
  \textbf{Gunnar Rätsch}\,$^{2}$ \qquad \quad
  \textbf{Guy Tennenholtz}\,$^{4}$ \\ 
  \\
$^{1}$\,ETH AI Center \qquad $^{2}$\,Department of Computer Science, ETH Zürich \\
  $^{3}$\,Max Planck Institute for Intelligent Systems, Tübingen\\
  $^{4}$\,Google Research, Mountain View \\
  \texttt{alizee.pace@ai.ethz.ch} \\
}

\begin{document}

\maketitle

\begin{abstract}
A prominent challenge of offline reinforcement learning (RL) is the issue of hidden confounding: unobserved variables may influence both the actions taken by the agent and the observed outcomes. Hidden confounding can compromise the validity of any causal conclusion drawn from data and presents a major obstacle to effective offline RL. In the present paper, we tackle the problem of hidden confounding in the nonidentifiable setting. We propose a definition of uncertainty due to hidden confounding bias, termed delphic uncertainty, which uses variation over world models compatible with the observations, and differentiate it from the well-known epistemic and aleatoric uncertainties. We derive a practical method for estimating the three types of uncertainties, and construct a pessimistic offline RL algorithm to account for them. Our method does not assume identifiability of the unobserved confounders, and attempts to reduce the amount of confounding bias. We demonstrate through extensive experiments and ablations the efficacy of our approach on a sepsis management benchmark, as well as on electronic health records. Our results suggest that nonidentifiable hidden confounding bias can be mitigated to improve offline RL solutions in practice.
\end{abstract}

\section{Introduction}

Large observational datasets for decision-making open the possibility of learning expert policies with minimal environment interaction. This holds promise for contexts where exploration is impractical, unethical or even impossible, such as optimising marketing, educational or clinical decisions based on relevant historical datasets \citep{thomas2017predictive,gottesman2018evaluating,singla2021reinforcement}. Recent years have thus seen the emergence of offline reinforcement learning (RL) literature \citep{Levine2020}, which proposes to adapt RL methods to overcome estimation biases induced by learning from finite, fully offline data. 

Aside from estimation biases, confounding variables are common in offline data \citep{gottesman2018evaluating}. The problem of hidden confounding, where outcome and decisions are both dependent on an unobserved factor, is widely overlooked in many of the concurrent offline RL methods. Nevertheless, it may induce significant errors, even for the simplest of bandit problems, and is especially aggravated in the sequential setting \citep{chakraborty2014dynamic,Zhang2019,Tennenholtz2022}. Hidden confounding exists in numerous applications. In autonomous driving, for example, the observational policy may behave according to unobserved factors (e.g. road conditions \citep{DeHaan2019}), which also affect environment dynamics and rewards. Alternatively, in the medical context, unrecorded patient state information such as socio-economic factors or visual appearance may have been taken into account by the acting physician \citep{gottesman2018evaluating}. 

In this work, we focus on \emph{nonidentifiable} hidden confounding in offline RL. While prior work has mostly addressed the problem in the identifiable setup \citep{lu2018deconfounding, Zhang2020DTR, kumor2021sequential, wang2021provably}, we show that significant improvement in policy learning can be achieved even in the realistic nonidentifiable setting. We propose an approach to estimate uncertainty due to confounding bias and to account for the degree of confoundedness while learning. In turn, this leads to improved downstream performance for offline learning algorithms. %

Our main contributions are as follows. (1) To the best of our knowledge, we are the first to address \emph{nonidentifiable} confounding bias in \emph{deep offline RL}.
(2) We achieve this by introducing a novel uncertainty quantification method from observational data, which we term delphic uncertainty.\linebreak  (3) We propose an offline RL algorithm that leverages this uncertainty to obtain confounding-averse policies, and (4) we demonstrate its performance on both synthetic and real-world medical data.

\section{Preliminaries}
\label{sec: preliminaries}

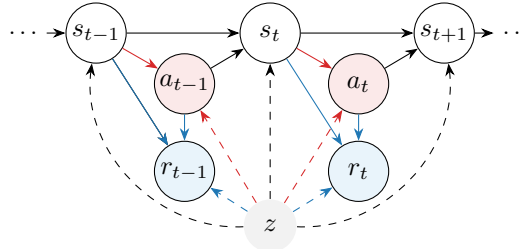
\begin{wrapfigure}[14]{r}{0.5\textwidth}
\vspace{-0.5cm}
\centering
\vspace{-1.5em}
\begin{tikzpicture}[auto,
    vertex/.style={minimum size=0.8cm,draw,circle,text width=0.8mm}, %
    vertex2/.style={circle,minimum size=0.7cm,fill=light-gray},%
    ]
        \node[vertex,label=center:$s_{t-1}$] (st-1) {};
        \node[vertex,right=1.5cm of st-1,label=center:$s_t$] (st) {}; %
        \node[vertex,right=1.5cm of st, label=center:$s_{t+1}$] (st+1) {};
        \node[right=0.25cm of st+1] (st+2) {\footnotesize $\ldots$};
        \node[left=0.25cm of st-1] (st-2) {\footnotesize $\ldots$};
        \node[vertex,below right=0.1cm and 0.6cm of st-1, label=center:$a_{t-1}$,fill=light-red] (at-1) {};
        \node[vertex,below right=0.1cm and 0.6cm of st, label=center:$a_t$,fill=light-red] (at) {};
        \node[vertex,below=0.35cm of at, label=center:$r_t$,fill=light-blue] (rt) {};
        \node[vertex,below=0.35cm of at-1, label=center:$r_{t-1}$,fill=light-blue] (rt-1) {};
        \node[vertex2,below=1.8cm of st] (x) {$z$};
        
        \path[-{Stealth[]},color=TabRed]
                (x) edge [dashed] (at)  
                (st) edge (at)
                (st-1) edge (at-1)
                (x) edge  [dashed] (at-1)
                  ;
         \path[-{Stealth[]}]
          (st) edge (st+1) 
          (at) edge (st+1)
          
          (st+1) edge (st+2)
          (st-1) edge (st)
          (st-1) edge (rt-1)
          (st-2) edge (st-1)
          (at-1) edge (st)
          (x) edge [dashed] (st)  
          (x) edge [dashed, bend right=50] (st+1)
          (x) edge [dashed, bend left=50] (st-1)
          ;

          \path[-{Stealth[]}, color=TabBlue]
          (st) edge (rt) 
          (at) edge (rt)
          (st-1) edge (rt-1)
          (at-1) edge (rt-1)
          (x) edge [dashed] (rt)
          (x) edge [dashed] (rt-1)
          ;

\end{tikzpicture}
\caption{\textbf{Contextual MDP.} Black arrows show the transition dynamics, blues ones the \textcolor{TabBlue}{reward function}, and red ones the \textcolor{TabRed}{policy}. Confounding arises when both behavioural policy $\behav$ and environment returns depend on hidden context variable $z$ (dashed lines).} 
\label{fig:cmdp}
\end{wrapfigure}

We consider the contextual Markov Decision Process (MDP) \citep{Hallak2015}, defined by the tuple ${\M = (\Sspace,  \Zspace, \Aspace, T, r, \rho_0, \nu, \gamma)}$, where $\Sspace$ is the state space, $\Aspace$ is the action space, $\Zspace$ is the context space, $T:\Sspace \times \Zspace \times \Aspace \rightarrow \Delta\Sspace$ is the transition function, $r:\Sspace \times \Zspace \times \Aspace \rightarrow [0, 1]$ is the reward function, and $\gamma \in [0, 1)$ is the discount factor. We assume an initial state distribution $\rho_0: \Zspace  \rightarrow \Delta\Sspace$ and a context distribution $\nu$, such that each interaction episode has a fixed context $z \sim \nu$, which may or may not be accessible to the agent, and the environment initialises at state $s_0 \sim \rho_0( \wc |z)$. At time $t$, the environment is at state $s_t \in \Sspace$ and an agent selects an action $a_t \in \Aspace$. The agents receives a reward $r_t = r(s_t, a_t, z)$ and the
environment then transitions to state $s_{t+1} \sim T( \wc | s_t, a_t, z)$. A causal graph of the process is depicted in \Cref{fig:cmdp}.

We define a \textit{context-aware} policy $\pi$ as a mapping $\pi: \Sspace \times \Zspace \rightarrow \Delta\Aspace$, such that $\pi( a |s, z)$ is the probability of taking action $a$ in state $s$ and context $z$. Likewise, we define a \textit{context-independent} policy as $\tilde{\pi}: \Sspace \rightarrow \Delta\Aspace$. We denote the set of all such policies as $\Pi$ and $\tilde{\Pi}$, respectively.\footnote{One can also consider history-dependent policies. Nevertheless, Markov policies sufficiently illustrate the challenges of our task, and can be easily generalised to history-dependent ones.}

We assume access to a dataset of $N$ trajectories $\mathcal{D} =  \{ \tau^i \}_{i=1}^{N}$, where the sequences ${\tau^i = ({s}_0^i, {a}_0^i, r_0^i, \ldots {s}_{H}^i, {a}_{H}^i, r_H^i)}$ are trajectories induced by an unknown, context-aware behavioural policy $\behav \in \Pi$ such that $a_t^i \sim \behav (\wc |s_t^i, z^i)$. The decision-making context $z^i$ for each trajectory is \textit{not} included in the observational dataset. In the following, we drop index $i$ unless explicitly needed.

Finally, we define the offline RL task with hidden confounding, which consists of finding an optimal context-independent policy $\tilde{\pi}^* \in \tilde{\Pi}$ -- one which maximises the expected discounted returns. Specifically, we define the state-action value function of a policy $\tilde{\pi} \in \tilde{\Pi}$ by $Q^{\tilde{\pi}}(s,a) = \Expt_{\tilde{\pi}, z \sim \nu}\left[ \sum_{t=0}^{\infty} \gamma^t r(s_t, a_t, z) \vert s_0 = s, a_0 = a\right]$, where $\Expt_{\tilde{\pi}}$ denotes the expectation induced by following policy $\tilde{\pi} \in \tilde{\Pi}$. We also define the value of $\tilde{\pi}$ by $V^{\tilde{\pi}}(s) = \Expt_{a \sim \tilde{\pi}} \left[ Q^{\tilde{\pi}}(s,a) \right]$. An optimal policy is then defined by
$
\tilde{\pi}^*(\wc|s) = \argmax_{\tilde{\pi} \in \tilde{\Pi}} \left[ V^{\tilde{\pi}}(s) \right].
$

\section{Sources of Error in Offline RL} \label{sec:sources of error}

Optimising a policy from observational data is prone to various sources of error, which many RL works propose to decompose, estimate, and bound \citep{Levine2020,tennenholtz2022uncertainty}. First, the process is prone to statistical error in correctly estimating a value model from the observed data \citep{jin2021pessimism}. Inherent stochasticity in the environment (aleatoric uncertainty) can result in imprecise models, whereas finite data quantities (epistemic uncertainty) can lead to poor model approximation.

When learning from observational data, improper handling of estimation errors causes covariate shift and overestimation problems, as evident in behaviour cloning \citep{Ross2011} and offline RL \citep{kumar2020conservative}. Such errors can be reduced through access to larger data quantities or online interactions with the true environment at training time. Offline RL approaches typically mitigate these errors through pessimism, penalising areas where error is expected to be large \citep{Levine2020, jin2021pessimism}. 

Another source of error, which is often overlooked in the RL literature, is structural bias. Independent of data quantity, such a bias can occur when the state-action space coverage is incomplete \citep{ueharapessimistic}, or when the expressivity of the model class considered is inappropriate \citep{lu2018nondelusional}. Our work considers confounding bias -- a critical type of structural bias, evident in a vast number of applications~\citep{DeHaan2019,kallus2018confounding,kallus2020confounding,gottesman2018evaluating}. This bias can arise when the data-generating policy relies on unobserved factors that also affect downstream transitions and/or rewards \citep{Tennenholtz2022}.

\paragraph{Confounding Bias.}

Confounding bias is a critical source of error in offline RL, which is often disregarded despite many data collection environments being prone to its occurrence \citep{kallus2018confounding}. This source of error arises when the observational policy depends on unobserved factors which affect the chosen action and the reward or transition function. To better understand how confounding bias may affect offline RL algorithms, consider the process detailed in \Cref{sec: preliminaries} and depicted in Figure~\ref{fig:cmdp}. The offline data was generated by sampling trajectories from the behavioural policy distribution $\tau \sim P_{\behav}(\tau)$, which is marginalised over $\nu(z)$ and factorises as follows:\begin{align}
     P_{\behav}(\tau) 
     = 
     \Expt_{\textcolor{red}{z} \sim \nu} \brk[s]*{ 
     \rho_0(s_0|\textcolor{red}{z}) \prod_{t=0}^H \behav(a_t|s_t,\textcolor{red}{z}) \: %
     P_r(r_t|s_t, a_t, \textcolor{red}{z}) \:
     T(s_{t+1}|s_{t},  a_{t},  \textcolor{red}{z})
     },
     \label{eq:prob_traj}
\end{align} 
where $P_r$ is the probability of sampling reward $r_t$ from $r(s_t,a_t,z)$.

Any offline reinforcement learning objective can be written as an expectation over this trajectory distribution \citep{Levine2020}. Confounding arises when one learns models on trajectories following $P_{\behav}(\tau)$, but estimates the value of policies $\pi$ that \textit{change} the probability of taking an action $a$ in a given state and context $(s,z)$ -- as is necessarily the case when considering context-independent policies. Since all model terms in \Cref{eq:prob_traj} are unknown and nonidentifiable due to their dependence on $z$, there may exist several ``worlds" that could induce the same observational distribution $P_{\behav}(\tau)$. This is known as the ``identifiability problem" in the causal inference literature \citep{kallus2018confounding,Namkoong2020}, which has been studied extensively, providing methods for analyzing when counterfactual estimates can be obtained. Particularly, without additional assumptions about the causal structure of the environment -- such as using environment interventions \citep{lu2018deconfounding,Zhang2020DTR} or the existence of observable back- or front-door variables  \citep{kumor2021sequential, wang2021provably} -- the context $z$ acting as confounder is nonidentifiable and cannot be estimated.  Below, we illustrate through a simple example, how two equally plausible models can correctly construct the same observational distribution, yet induce two different values for another policy.

\begin{figure}[t!]
     \centering
     \begin{subfigure}[b]{0.23\textwidth}
         \centering \includegraphics[trim=0.2cm 0 0 0, clip,height=4.2cm]{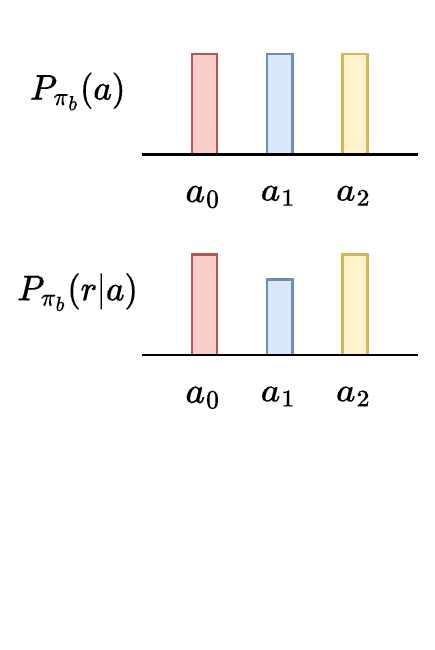}
        \caption{Observational data} \label{fig:poc_data}
     \end{subfigure} \hfill %
     \begin{subfigure}[b]{0.25\textwidth}
         \centering \hspace{-1em}\includegraphics[trim=0.1cm 0 0.3cm 0, clip,height=4.2cm]{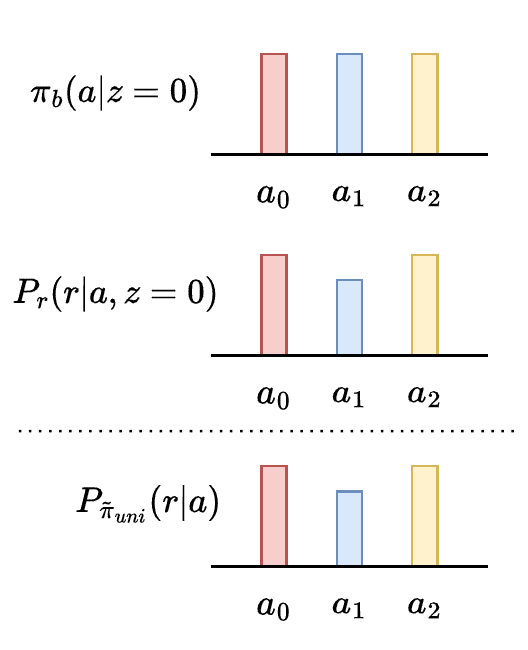}
         \caption{World 1: $\nu(0)=1$.} \label{fig:poc_wm1}  
     \end{subfigure} %
     \begin{subfigure}[b]{0.46\textwidth}
         \raggedleft \includegraphics[trim=0 0 0.1cm 0 , clip,height=4.2cm]{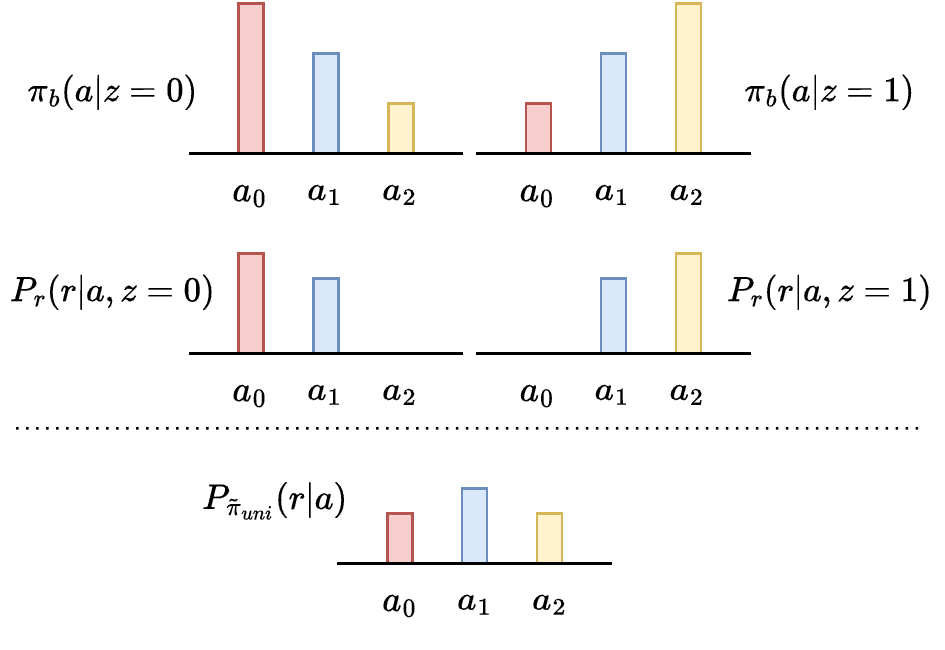}%
         \caption{World 2: $\nu(0)=0.5$.} \label{fig:poc_wm2} 
     \end{subfigure} 
    \caption{\footnotesize \textbf{Confounding Bias Example.} World 1 and 2 are two models for the binary confounding variable that are compatible with the marginalised observational bandit data in (a), composing models for $\nu(z)$, $\behav(a|z)$ and $P(r|a,z)$. Under an alternative policy, such as a context-independent uniform policy $\tilde{\pi}_{uni} (\cdot) = 1/|\Aspace|$, these two worlds give different values to each action.}
    \label{fig:banditexample}
\end{figure}

\paragraph{An Illustrative Example.} Suppose access to the bandit data in Figure \ref{fig:poc_data}, induced by an unknown context-dependent policy $\behav$ with marginal distribution $P_{\behav}(a, r)$. Assume no access to the episode context $z$ in the data. Simplifying \Cref{eq:prob_traj} to this setup, we obtain:
\begin{align*}
    P_{\behav}(a, r)
    =
    \Expt_{z \sim \nu} [\behav(a|z) P_r(r|a,z)].
\end{align*}
We can therefore change $\nu$, $\behav$, and $P_r$ to induce the same marginalised distribution $P_{\behav}$, with a significant difference in reward for a counterfactual policy. Indeed, in Figures \ref{fig:poc_wm1} and \ref{fig:poc_wm2} we show how different models that are compatible with the observational quantities can result in substantially different reward estimates for a different policy. Particularly, in World 1 (\Cref{fig:poc_wm1}), we assume a deterministic singleton context, with a corresponding uniform behavioural policy, whereas in World 2 (\Cref{fig:poc_wm2}) we assume two contexts with uniform distribution, and a behavioural policy which changes its distribution w.r.t. the sampled context. In both of these worlds, the observational distribution $P_{\behav}(a, r)$ remains the same. Nevertheless, calculating the reward of the uniform policy $\tilde{\pi}_{uni} (\cdot) = 1/|\Aspace|$ results in different reward distributions. Moreover, the optimal actions in World 1 and World 2 are different. 

Without explicit access to the ground-truth context or a proxy thereof (in the identifiable context), modelling an alternative policy to the privileged data-generating one will therefore be prone to spurious correlations and estimation biases. 

In the next section, we propose to address the general \textit{nonidentifiable} confounding problem in offline RL by estimating the amount of confounding error within the observational dataset and correcting for it during learning. Importantly, this source of error cannot be captured by epistemic or aleatoric uncertainty quantification methods, as discussed next.

\section{Measuring Confounding Bias through Delphic Uncertainty}

In this section, we formulate a method for estimating uncertainty arising from confounding bias in offline RL, which we term \textit{delphic} uncertainty\footnote{The word ``delphic" characterises quantities that are ambiguous and opaque, relating to the hidden confounding variables and their elusive effect on model predictions.}. While aleatoric and epistemic uncertainty can be expressed as probability distributions over model outputs and parameters, respectively \citep{hullermeier2021aleatoric}, delphic uncertainty is a distribution over counterfactual values. We propose a general approach to decouple aleatoric, epistemic, and delphic uncertainties, which we later leverage to overcome confounding bias in \Cref{sec:method}.

To introduce delphic uncertainty we first define a set of worlds compatible with the marginalised data distribution $P_{\behav}(\tau)$. 
\begin{definition}
\label{def: compatible world models}
    A compatible world for $P_{\behav}$ is a tuple $w=(\Zspace_w, \nu_w, \rho_{0,w}, P_{r,w}, T_w, \pi_{b,w})$ which satisfies $P_{\behav}(\tau) = \Expt_{z \sim \nu_w} \brk[s]*{\rho_{0,w}(s_0|z)\prod_{t=0}^H \pi_{b,w}(a_t|s_t,z) P_{r,w}(r_t|s_t, a_t, z) T_w(s_{t+1}|s_{t}, a_{t}, z)}$ for any trajectory $\tau = (s_0, a_0, r_0, \hdots, s_H, a_H, r_H)$. We denote by $\W$ the set of all compatible worlds.
\end{definition}

We focus on uncertainty estimates of value functions. Let $w \in \W$ (i.e., $w$ is some compatible world for $P_{\behav}$). We use $\theta_w$ to denote the parameters of a Q-value function in world $w$. For a fixed $w$ and $\theta_w$, we assume each value model $Q_{\theta_w}$ is defined by some stochastic model, e.g., a normal distribution $Q_{\theta_w} | \theta_w, w \sim \mathcal{N}\brk*{\mu_{\theta_w}, \sigma_{\theta_w}^2}$. Indeed, here $\sigma_{\theta_w}$ accounts for \textbf{aleatoric uncertainty}, capturing the intrinsic stochasticity of the environment \citep{kendall2017uncertainties}. Additional statistical uncertainty arises from the distribution over model parameters $\theta_w$ in the fixed world $w \in \W$. Starting from a prior over $\theta_w$, evidence from the data leads to a posterior estimate over the correct model parameters $P(\theta_w| \dataset)$, which captures \textbf{epistemic uncertainty} \citep{hullermeier2021aleatoric}. We refer the interested reader to \Cref{appendix:related_works_uncertainty} for an overview of statistical uncertainty estimation methods. %

We are now ready to define the uncertainty induced by confounding variables, which we term \textbf{delphic uncertainty}. To do this, we leverage \Cref{def: compatible world models} and define delphic uncertainty by varying over compatible world models. Based on the law of total variance \citep{weiss2006course} and following on prior work separating epistemic and aleatoric uncertainty \citep{kendall2017uncertainties}, we can decompose the variance in the value function estimate between the three types of uncertainties. Particularly, let $w$ be a compatible world for $P_{\behav}$, and let $P_{w} \mapsto \Delta\W$ be some distribution over worlds in $\W$. We have the following result. Its proof, based on the law of total variance \citep{weiss2006course}, is given in \Cref{appendix:proofs}.

\begin{theorem}[Variance Decomposition]
\label{thm:var_decomp}
For any $\pi \in \Pi$, we have
\begin{align*}  \resizebox{\hsize}{!}{
$\Var (Q^{\pi}_{\theta_w}) 
= 
\Expt_{w} \left[\underbrace{\Expt_{\theta_w} \left[\Var(Q^{\pi}_{\theta_w} | \theta_w, w) | w \right]}_\text{\normalsize{aleatoric uncertainty}}  
+ 
\underbrace{\Var_{\theta_w}\left(\Expt[Q^{\pi}_{\theta_w} |  \theta_w, w] | w\right)}_\text{\normalsize{epistemic uncertainty}}  \right] 
+
\underbrace{\Var_{w} (\Expt_{\theta_w}[\Expt[Q^{\pi}_{\theta_w}| \theta_w, w]|w ])}_{\text{\normalsize{delphic uncertainty}}}.$
}
\end{align*}
\end{theorem}

To gain further intuition of this result, consider the case of normal distributions. We can rewrite \Cref{thm:var_decomp} as:
\begin{align}
 \Var (Q^{\pi}_{\theta_w}) 
 =
 \Expt_{w} 
 \left[
 \Expt_{\theta_w} [\sigma_{\theta_w} | w]^2
 + %
 \Var_{\theta_w}(\mu_{\theta_w} | w) + \Var_{\theta_w}(\sigma_{\theta_w}| w)
 \right]
 +
\Var_{w} (\Expt_{\theta_w}[\mu_{\theta_w}|w ]) 
\label{eq:total_var}
\end{align}
The first three terms, calculated by the square average of predicted standard deviations and the variance of the predicted means and standard deviations, correspond to aleatoric and epistemic uncertainties, whereas the final term, calculated by the variation over compatible world models, corresponds to delphic uncertainty. Indeed, the latter form of uncertainty cannot be diminished, even in deterministic environments and infinite data: as $\abs{\dataset} \to \infty$ (no epistemic uncertainty), and $\sigma_{\theta_w} \to 0$ (no aleatoric uncertainty), the delphic uncertainty remains. We refer the reader to \Cref{appendix:asymptotic} for further discussion.

\section{Offline RL Under Delphic Uncertainty} 
\label{sec:method} 

In the previous section, we defined delphic uncertainty through variation over compatible world models. In this section, we propose a method to measure delphic uncertainty in practice. We then leverage our uncertainty estimate within an offline reinforcement learning framework and demonstrate its ability to mitigate confounding bias. 

Following the estimation approach outlined in \Cref{thm:var_decomp}, delphic uncertainty can be measured through the disagreement within value functions for a given policy, under different worlds $w$ compatible with the observational distribution. %

\begin{wrapfigure}[10]{r}{0.5\textwidth}
    \centering \vspace{-5em}
    \includegraphics[width=\textwidth]{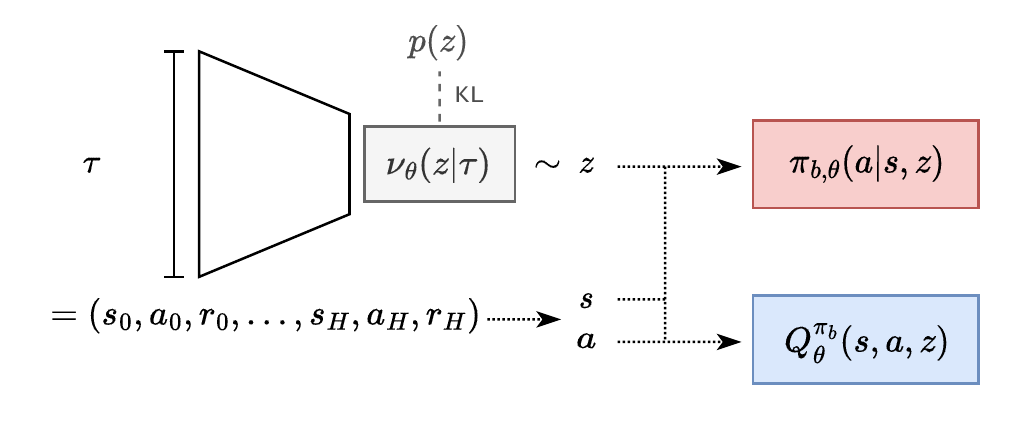} %
    \caption{\footnotesize \textbf{Individual world model architecture} $w = ( \nu_\theta, \pi_{b,\theta}, Q^{\behav}_\theta)$, under a prior $p(z)$ for the confounder distribution. Multiple worlds are trained and their variance in estimating $Q ^{\pi}$ is taken as delphic uncertainty.}
    \label{fig:architecture}
\end{wrapfigure}

\subsection{Measuring Delphic Uncertainty}

\paragraph{Modelling Compatible Worlds.} The first practical step for evaluating delphic uncertainty is the definition of compatible world models. While one could theoretically consider all possible world models in \Cref{def: compatible world models}, we found that, in practice, varying over a subset of compatible models was enough to show improved offline RL efficiency.

A compatible world $w \in \W$ must capture key relationships from the observational data.  \Cref{fig:architecture} depicts our proposed approach. Our model, parameterised by $\theta$, consists of a confounder prior, a behaviour policy, and a value function estimator. During training, a trajectory $\tau \sim \dataset$ is mapped to a latent distribution $\nu_\theta (z|\tau)$, from which the policy $\pi_{b,\theta}$ and value $Q^{\behav}_\theta$ are estimated. Estimates are trained by draws from state-action pairs $(s,a) \sim \tau$ and a sampled $z$. 

More specifically, we train compatible world models through variational inference, using the posterior $\nu_\theta(z|\tau)$ and prior $p(z)$. For $\tau \sim \dataset$, the model is trained by maximising the Evidence Lower BOund (ELBO, \citet{kingma2014}): \begin{equation*}
    \Expt_{(s,a) \sim \tau; \: z \sim \nu_\theta(z|\tau)} 
    \brk[s]!{
     \log Q^{\behav}_\theta(s,a,z) + \alpha \log \pi_{b,\theta}(a|s,z)
     }
     - \beta D_{KL} \brk!{ \nu_\theta (z|\tau) \,\big\Vert\, p(z) }
\end{equation*}
where $\{\alpha, \beta\}$ are hyperparameters \citep{betaVAE} and $D_{KL}$ is the Kullback-Leibler divergence between two distributions. Sampling from $\nu_\theta$ is achieved through the reparametrisation trick \citep{kingma2014}. Optimal parameters for $\{ \nu_\theta, \pi_{b,\theta}, Q^{\behav}_\theta\}$ are obtained by maximising the objective over $\dataset$. Once a compatible world model is trained, the value function of a policy $\pi$ can be estimated over one step using importance sampling and marginalising over $z$, i.e., $Q^{\pi} (s,a) =\Expt_{\tau \sim \dataset}  \Expt_{z \sim \nu_\theta(z|\tau)} \brk[s]*{\frac{\pi(a|s)}{\pi_{b,\theta}(a|s,z)} Q^{\behav}_\theta (s,a,z) } $. We discuss alternative approaches to estimating counterfactual quantities in \Cref{appendix:implementation}.

\paragraph{Counterfactual Variation Across Worlds.} We approximate $\W$ by a set of $W$ compatible worlds $\{ \nu_{\theta_w}, \pi_{b,\theta_w}, Q^{\behav}_{\theta_w}\}_{w=1}^W$ on the observational training dataset $\dataset$, each trained using different priors and model architectures. Following our definition of confounding bias (\Cref{thm:var_decomp}), we measure delphic uncertainty through the variance in $Q_{\theta_w} ^{\pi} (s,a)$ across worlds. That is, delphic uncertainty for policy $\pi$ at state-action $(s,a)$ is defined by $u_d ^{\pi}(s,a) = \Var_w ( Q_{\theta_w} ^{\pi} (s,a) )$. When no confounding exists, all models in $\W$ should identify similar $\nu, Q^{\behav}$ and $\behav$ (up to epistemic uncertainty), returning a similar value of $Q^{\pi}$. On the other hand, confounding with ambiguous returns would lead to different values across world models. Epistemic and aleatoric uncertainty are separately captured by implementing each world model component as an ensemble of probabilistic models. We refer the reader to \Cref{appendix:implementation} for an exhaustive overview of the training procedure.

\subsection{Delphic ORL: Offline Reinforcement Learning with Delphic Uncertainty} \label{sec:pessimism}

Inspired by pessimistic approaches in offline RL \citep{Kumar2019, kumar2020conservative, fujimoto2019off, Levine2020, jin2021pessimism}, we propose to penalise the value of states and actions where delphic uncertainty is high, such that the learned policy is less likely to rely on spurious correlations between actions, states and rewards. This pessimistic approach, which enables the agent to account for and mitigate confounding bias when making decisions, is summarised in \Cref{alg:algo}.

      \begin{algorithm}[t!]
        \caption{Delphic Offline Reinforcement Learning} \label{alg:algo} 
        \SetKw{Input}{Input:}
        \Input{Observational dataset $\dataset$, Offline RL algorithm.} \\
        Learn compatible world models $\{\Zspace_w, \nu_w, \rho_{0,w}, P_{r,w}, T_w, \pi_{b,w}\}_{w\in \W}$ that all factorise to $P_{\behav}(\tau)$. \\ %
        Obtain counterfactual predictions $Q_w ^{\pi}$ for each $w \in \W$. \\
        Define local delphic uncertainty: $u_d^{\pi} (s,a) = \Var_w ( Q_w ^{\pi} (s,a) )$. \\
        Apply pessimism using $u_d$ in Offline RL algorithm (see \Cref{sec:pessimism}) \\    
      \end{algorithm}

In this paper, we incorporate pessimism with respect to delphic uncertainty by modifying the target $Q_{target}$ for the Bellman update in a model-free offline RL algorithm to $$Q_{target}'(s, a) = Q_{target}(s, a) - \lambda u_d^{\pi} (s, a),$$ where $\pi$ is the latest learned policy, $(s,a)$ is a tuple sampled for the update and hyperparameter $\lambda$ controls the penalty strength. We apply our penalty to Conservative Q-Learning~\citep{kumar2020conservative}, but this approach could also be implemented within any model-free offline RL algorithm -- which already induces pessimism with respect to epistemic uncertainty \citep{kumar2020conservative,fujimoto2021minimalist,Levine2020}. %

Note that various other methods can be adopted to drive pessimism against delphic uncertainty within existing offline RL algorithms \citep{kumar2020conservative, yu2020mopo, fujimoto2021minimalist}, depending on the task at hand. The above penalty can be subtracted from the reward function in model-based methods \citep{yu2020mopo}. The uncertainty measure can also be used to identify a subset of actions over which to optimise the policy, as demonstrated by \citet{fujimoto2019off}, or to weigh samples in the objective function, prioritising unconfounded data. We refer the reader to~\Cref{appendix:implementation} for implementation details, including on the aforementioned techniques.

\section{Experiments}

In this section, we study the benefits of our proposed delphic uncertainty estimation method and its application in offline RL. We validate two principal claims: (1) Our delphic uncertainty measure captures bias due to hidden confounders. (2) \Cref{alg:algo} leads to improved offline RL performance in both simulated and real-world confounded decision-making problems, compared to state-of-the-art but biased approaches. As baselines compatible with the discrete action spaces of environments studied here, we consider Conservative Q-Learning (CQL) \citep{kumar2020conservative}, Batch-Constrained Q-Learning (BCQ) \citep{fujimoto2019off} and behaviour cloning (BC) \citep{Bain1996}. Implementation and dataset details are provided in \Cref{appendix:implementation,appendix:expdetails} respectively. In the following, we measure and vary confounding strength through the dependence of the behavioural policy on the hidden confounders, $\Gamma = \max_{z,z' \in \Zspace} [{\behav(a|s,z)}/{\behav(a|s,z')}]$ \citep{rosenbaum2002}, where $z$ also affects the transition dynamics or reward function.

\subsection{Sepsis Simulation}

We explore a simulation of patient evolution in the intensive care unit adapted from \citet{Oberst2019}. The diabetic status of a patient, accessible to the near-optimal behavioural policy but absent from the observational dataset, acts as a hidden confounder $z$. 
\begin{figure}
     \centering
     \begin{subfigure}[b]{0.33\textwidth}
         \centering \includegraphics[height=3cm]{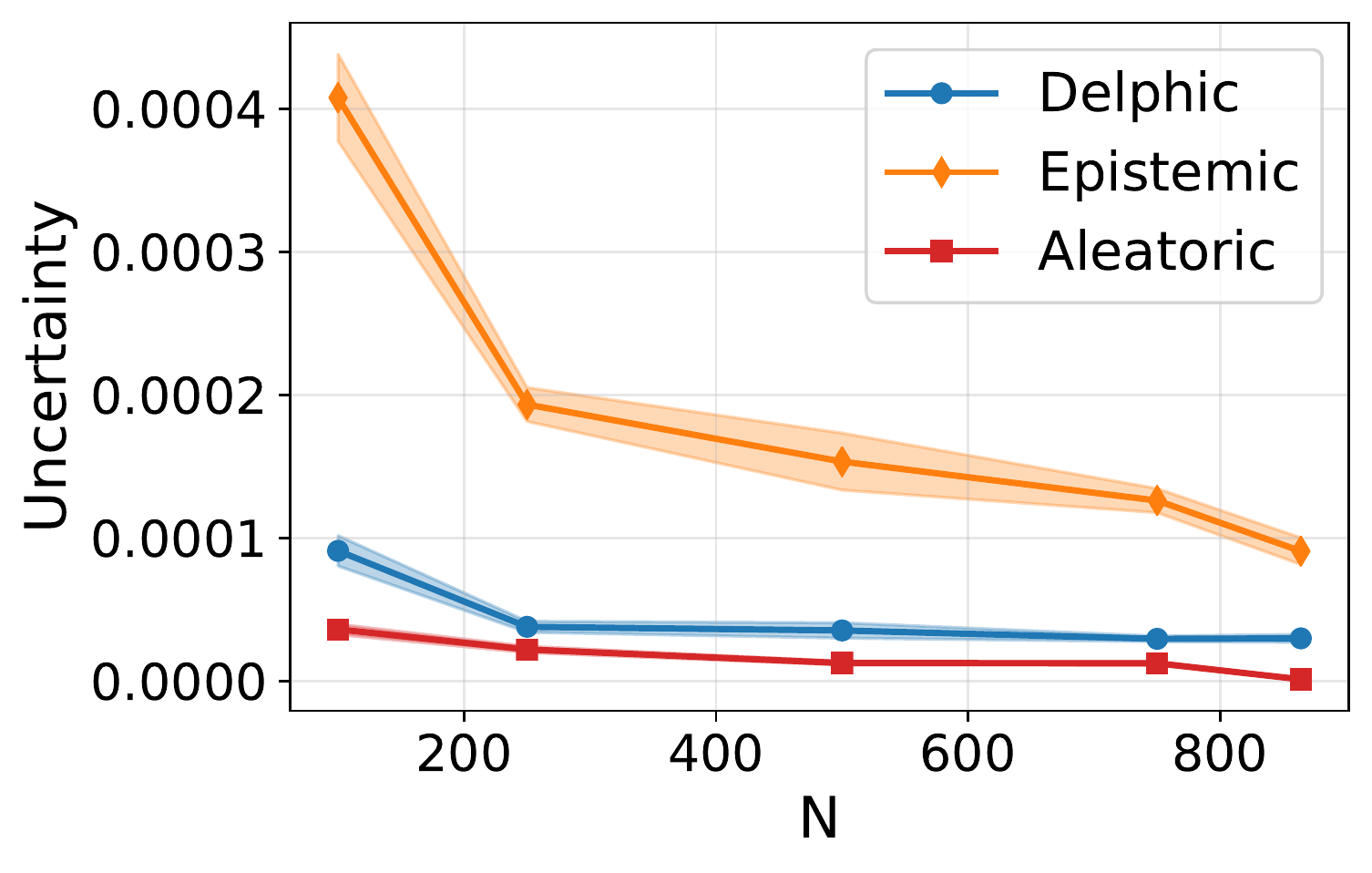}
        \caption{Number of trajectories $N$} \label{fig:u_vs_N_real}
     \end{subfigure} \hfill
      \begin{subfigure}[b]{0.31\textwidth}
         \centering \includegraphics[height=3cm]{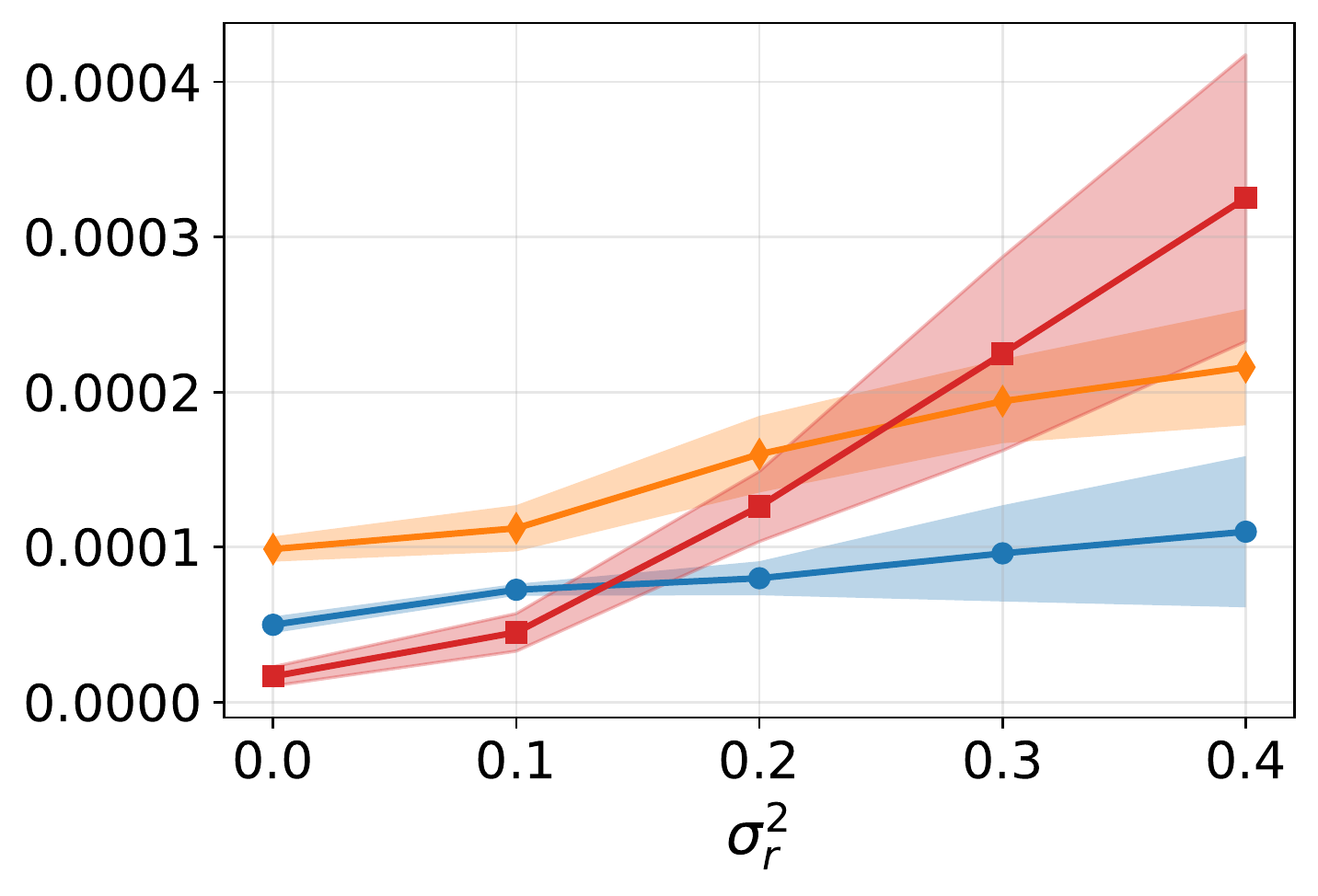}
         \caption{Environment stochasticity $\sigma^2_r$} \label{fig:u_vs_stocha_real}
     \end{subfigure}  \hfill
     \begin{subfigure}[b]{0.31\textwidth}
         \centering \includegraphics[height=3cm]{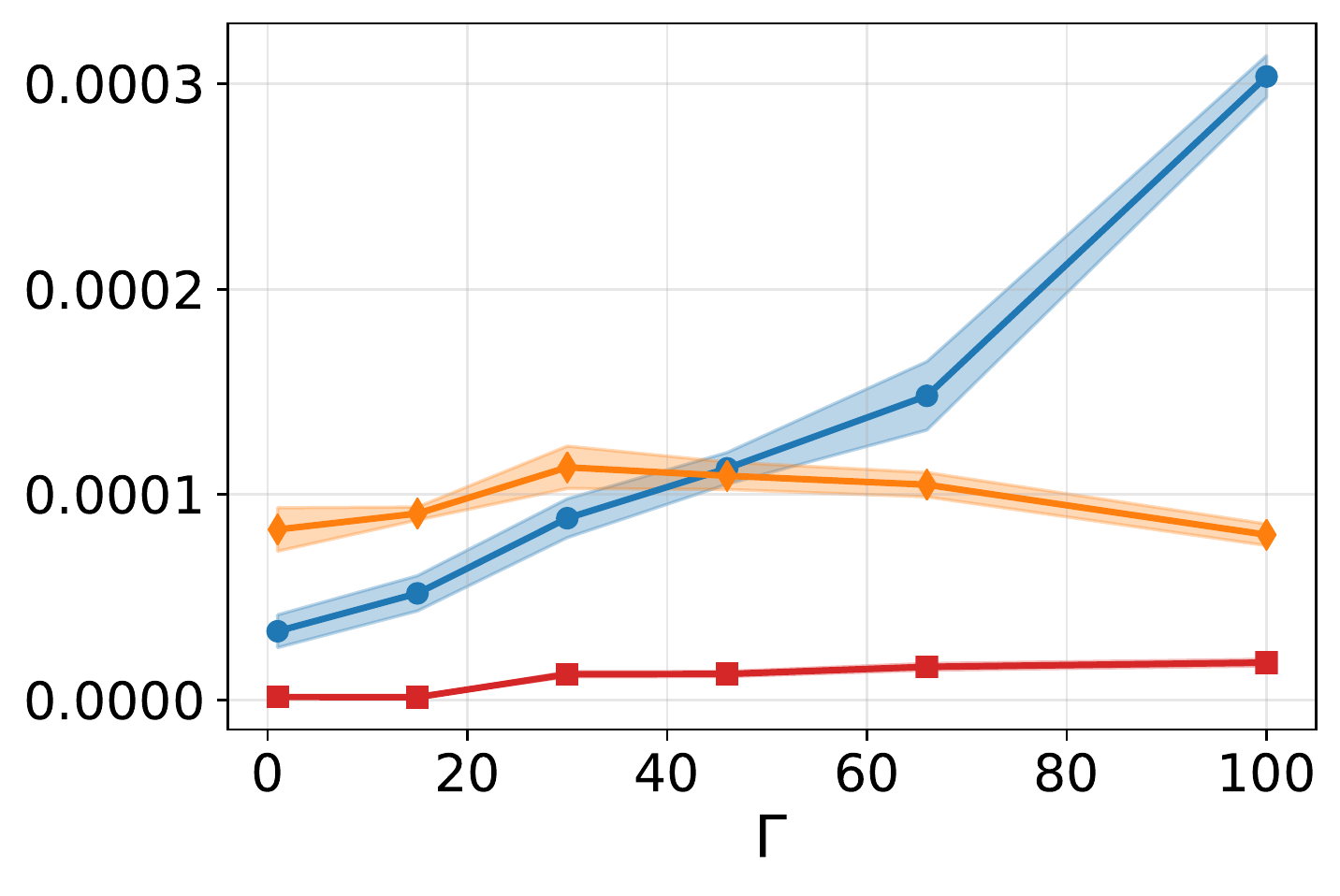}
         \caption{Confounding strength $\Gamma$} \label{fig:u_vs_conf_real}
     \end{subfigure} \\ 
    \caption{\footnotesize \textbf{Uncertainty measures as a function of data properties}, averaged over state-action pairs in the sepsis dataset. Epistemic uncertainty reduces most with more data, aleatoric uncertainty increases most with environment stochasticity (reward variance), and delphic uncertainty increases most with confounding strength.}
    \label{fig:u_vs_NZ}
\end{figure}

\paragraph{Uncertainty Measures.} First, we study the relationship between our uncertainty estimates and the decision-making setup. In Figure \ref{fig:u_vs_NZ}, we find that epistemic uncertainty reduces with greater data quantities and increases out of the training set distribution, whereas aleatoric uncertainty increases with environment stochasticity, in agreement with prior work~\citep{kendall2017uncertainties}. Our delphic uncertainty estimate, on the other hand, cannot be reduced with more data and increases with greater confounding. Moreover, we found that delphic uncertainty relates to meaningful regions of state-action space, as it is highest under vasopressor administration -- the only treatment for which patient evolution is confounded by the hidden diabetic status. We refer the reader to \Cref{appendix:extra_results} for an exhaustive overview and further experiments.

\begin{wrapfigure}[15]{r}{0.45\textwidth}
    \centering
    \vspace{-1.7em}
    \includegraphics[width=\textwidth]{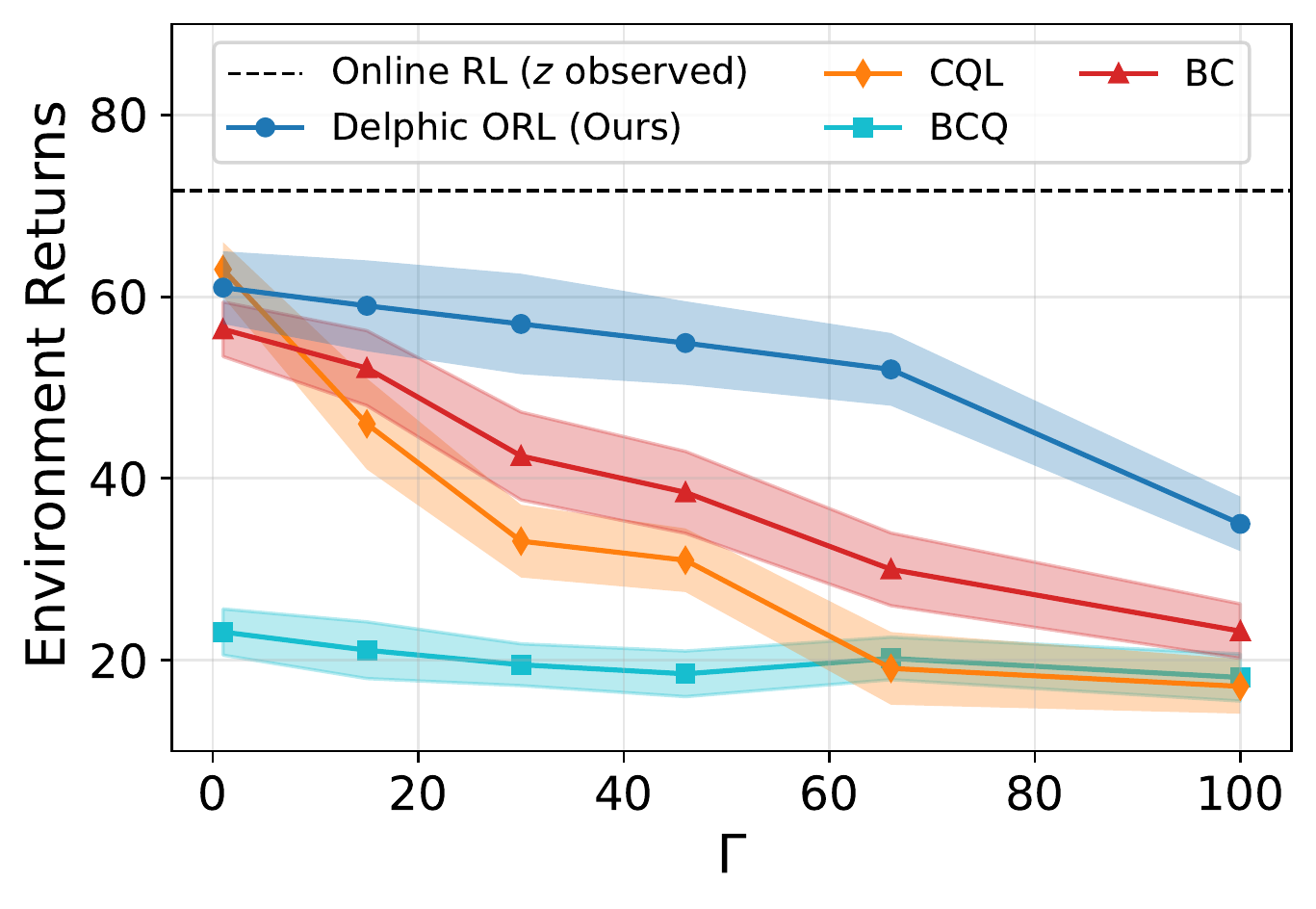} %
    \caption{\footnotesize \textbf{Performance Results} as a function of confounding strength $\Gamma$. Normalised environment returns (mean and shaded 95\% CIs) over 10 runs.}
    \label{fig:perf_vs_conf}
\end{wrapfigure} 

\paragraph{Offline RL Performance.} In \Cref{fig:perf_vs_conf}, we compare environment returns obtained through offline RL, imitation learning, and our proposed approach. Our results reveal the susceptibility of offline RL to confounding bias: the presence of unobserved factors $z$ that influence both the behaviour policy and transition dynamics leads to inaccurate value function estimates. Behaviour cloning appears to be less prone to this bias but still faces challenges in dealing with missing information in $z$, evidenced by the performance gap to the online policy in the unconfounded case ($\Gamma=1$), and with the distribution shift in observed histories \citep{ortega2021shaking}. In contrast, our approach to penalising delphic uncertainty leads to superior performance, especially as confounding strength increases. In \Cref{appendix:extra_results}, we also compare different approaches to implementing pessimism with respect to delphic uncertainty, as detailed in \Cref{sec:pessimism}, and provide an ablation over performance as a function of pessimism hyperparameter $\lambda$.

\subsection{Real-World Data}

We demonstrate the added value of our algorithm in optimising decision-making policies from real-world medical data. Our clinical policies are trained using a publicly available dataset of electronic health records, with over 33 thousand patient stays in intensive care and over 200 measured variables \citep{Hyland2020}. We consider the problem of optimising the treatment policy for vasopressor and fluid administration\footnote{These therapeutic agents are commonly given to overcome shock in intensive care \citep{benham2012cardiovascular}. Their administration strategy has already been studied as an RL task \citep{Raghu2017,gottesman2018evaluating}.}, and design the reward function to avoid states of circulatory failure. Significant information about patients' conditions is not available in the dataset, despite being critical to the treatment choices of attending physicians, such as socio-economic factors or medical history \citep{Yang2018}. For ease of evaluation, we introduce additional, artificial confounders by excluding them from the observational dataset, focusing on diagnostic indicator variables, age and weight ($\Gamma \in [1, 200]$). Disease severity is measured through the SOFA score system \citep{vincent1996sofa}.

\begin{figure}[tb]
\centering
     \begin{subfigure}[b]{0.31\textwidth}
         \centering \includegraphics[width=\textwidth]{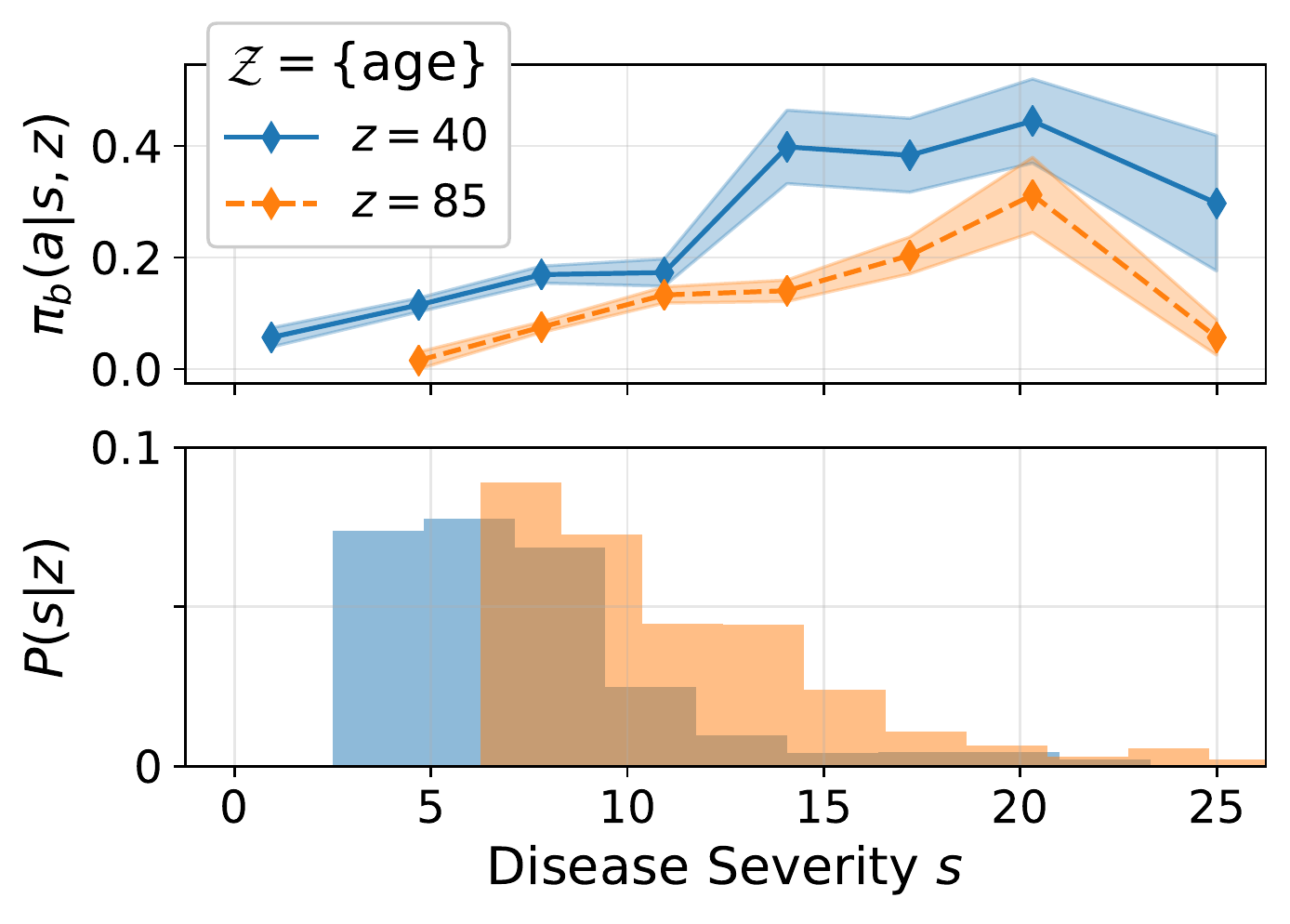}
         \caption{$\{$age$\}$, $\Gamma=48.2$}
     \end{subfigure} \hfill
      \begin{subfigure}[b]{0.31\textwidth}
         \centering \includegraphics[width=\textwidth]{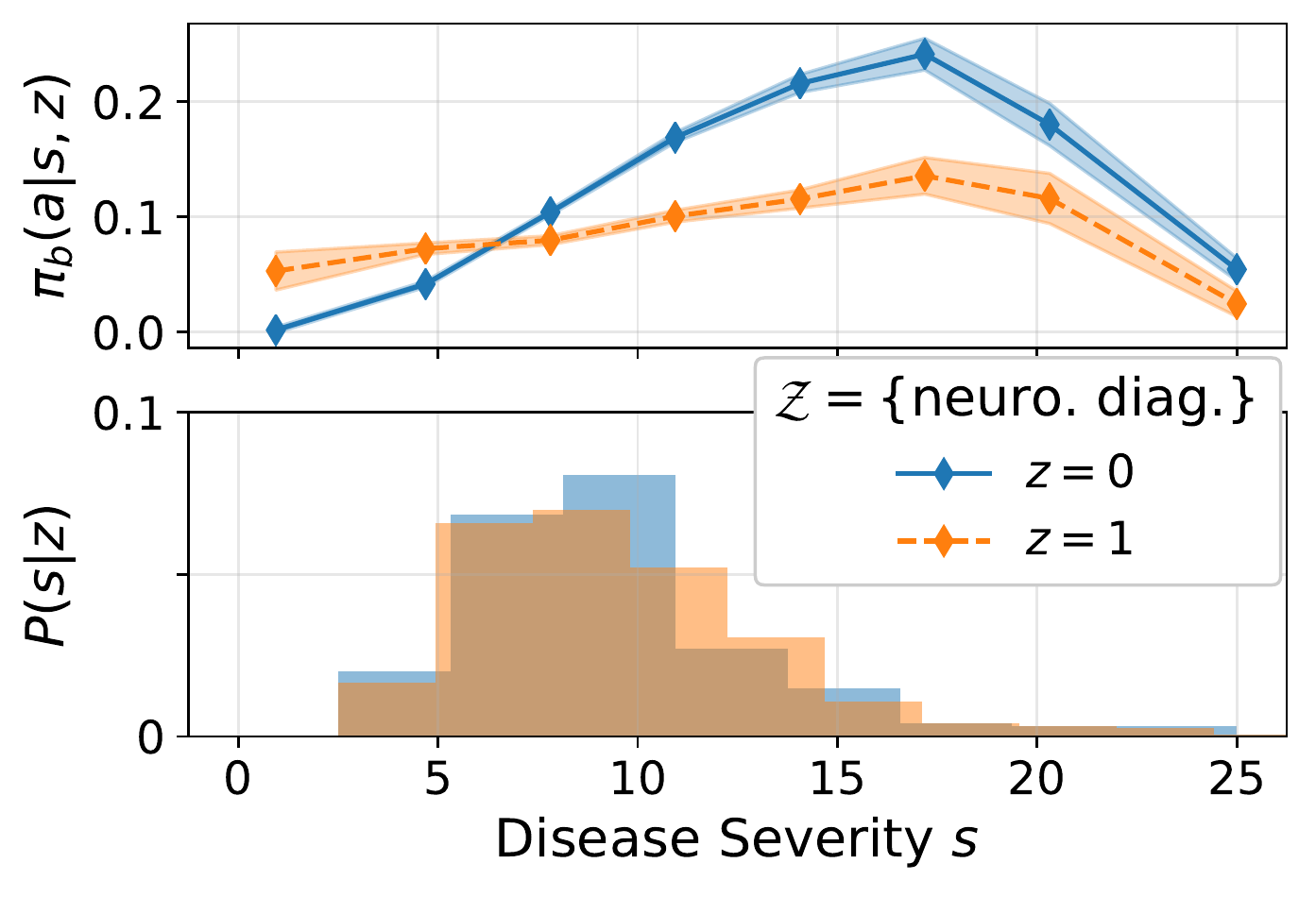}
        \caption{$\{$neuro. diag.$\}$, $\Gamma=29.1$}
     \end{subfigure} \hfill
     \begin{subfigure}[b]{0.36\textwidth}
         \centering \includegraphics[width=\textwidth]{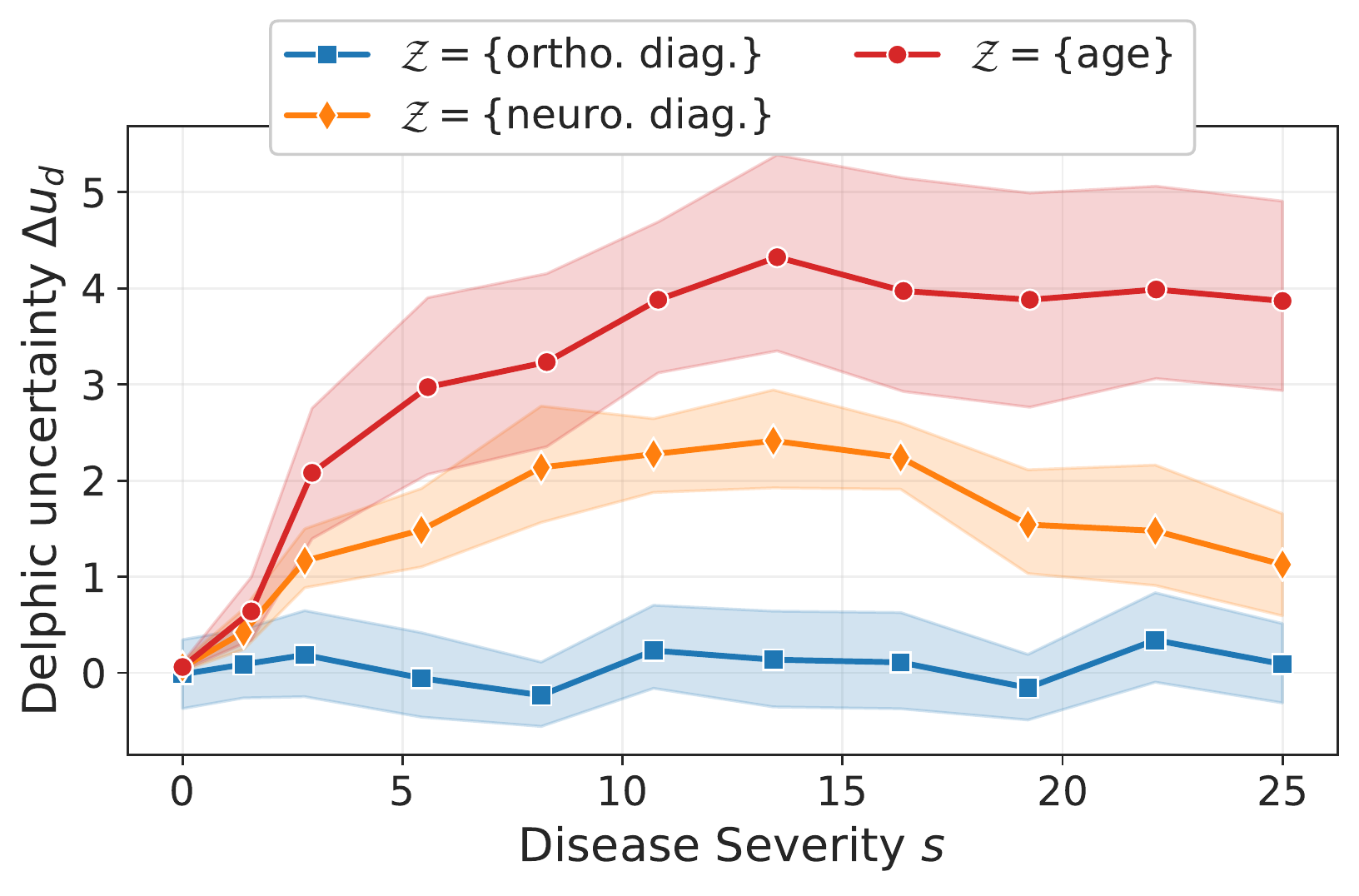}
         \caption{Uncertainty increase under $\Zspace$.} \label{fig:u_increase_vs_z}
     \end{subfigure}
    \caption{\footnotesize \textbf{Delphic uncertainty as a function of $(s,z)$} in real-world medical data, for $a=\{$vasopressors$\}$. In (a, b), we note the dependence of the behavioural policy $\behav$ (top) and/or state distribution $P(s)$ (bottom) on confounders $z$. In (c), delphic uncertainty increases most in confounded states and under factors with greater confounding strength, compared to orthopaedic diagnosis ($\Gamma=3.4$).}  
    \label{fig:confounding_ICU}
\end{figure}

\paragraph{Confounding in Medical Dataset.} As the aforementioned variables affect both the probability of treatment assignment and downstream patient evolution, they act as confounders over outcome models when excluded from the data. In Figure~\ref{fig:confounding_ICU}, we highlight how our delphic uncertainty measure captures confounded state-action pairs in concordance with the introduced confounders. Delphic uncertainty is generally highest for high disease severity, where important factors such as age or comorbidity may affect the choice of treatment intensity \citep{azoulay2009}. Indeed, delphic uncertainty increased to a greater extent under important confounders (e.g., age or patients' neurological diagnosis) than less critical factors (e.g., orthopaedic diagnosis).

\begin{figure}[tb]
    \centering \begin{floatrow}%
\capbtabbox{%
      \resizebox{0.56\textwidth}{!}{%
     \begin{tabular}{lccccc}
    \toprule
        Confounders $\Zspace$ & $\Gamma$ & BCQ & BC & CQL & Delphic ORL \\ \midrule
        All 14 & $\approx$200 & 54.6 $\pm$ 1.3 & 59.6 $\pm$ 0.8 & 59.3 $\pm$ 0.9 & \textbf{62.2} $\pm$ 1.0 \\ \midrule
        $\{$age$\}$ & 48.2 & 58.8 $\pm$ 0.8 & 64.7 $\pm$ 0.5 & 64.4 $\pm$ 0.8 & \textbf{66.5} $\pm$ 0.9 \\
        $\{$neuro. diag.$\}$ & 29.1 & 55.0 $\pm$ 1.3 & 61.8 $\pm$ 0.9 & 59.6 $\pm$ 1.7 & \textbf{65.7} $\pm$ 1.2 \\
        $\{$gastro. diag.$\}$ & 19.0 & 55.8 $\pm$ 0.8 & 60.9 $\pm$ 0.6 & 59.8 $\pm$ 0.6 & \textbf{63.3} $\pm$ 1.1 \\
        $\{$trauma$\}$ & 16.3 & 56.3 $\pm$ 0.8 & 63.2 $\pm$ 1.1 & 63.5 $\pm$ 0.7 & \textbf{65.7} $\pm$ 1.0 \\
        $\{$cardio. diag.$\}$ & 13.2 & 56.2 $\pm$ 1.0 & 60.6 $\pm$ 0.7 & 58.6 $\pm$ 0.9 & \textbf{62.7} $\pm$ 1.1 \\
        $\{$hemato. diag.$\}$ & 11.6 & 59.6 $\pm$ 0.9 & 63.2 $\pm$ 0.6 & 63.1 $\pm$ 0.7 & \textbf{65.3} $\pm$ 1.1 \\
        $\{$weight$\}$ & 8.3 & 60.1 $\pm$ 0.8 & \textbf{64.2} $\pm$ 1.0 & \textbf{65.4} $\pm$ 0.6 & \textbf{66.3} $\pm$ 0.9 \\
        $\{$sedation$\}$ & 6.8 & 61.2 $\pm$ 0.8 & \textbf{64.5} $\pm$ 0.6 & \textbf{64.8} $\pm$ 0.9 & \textbf{65.3} $\pm$ 1.2 \\
        $\{$endo. diag.$\}$ & 4.7 & 60.1 $\pm$ 1.1 & 63.1 $\pm$ 0.6 & \textbf{65.5} $\pm$ 0.8 & \textbf{65.7} $\pm$ 1.0 \\
        $\{$resp. diag.$\}$ & 4.4 & 61.6 $\pm$ 1.3 & \textbf{64.0} $\pm$ 0.9 & \textbf{65.9} $\pm$ 1.0 & \textbf{64.7} $\pm$ 1.0 \\
        $\{$ortho. diag.$\}$ & 3.4 & 62.3 $\pm$ 0.8 & \textbf{64.6} $\pm$ 0.6 & \textbf{65.8} $\pm$ 0.7 & \textbf{65.9} $\pm$ 1.0 \\
        $\{$surgical status$\}$ & 3.2 & 62.2 $\pm$ 1.1 & 64.3 $\pm$ 0.5 & \textbf{67.4} $\pm$ 0.7 & \textbf{66.8} $\pm$ 1.1 \\
        $\{$sepsis$\}$ & 2.8 & 60.3 $\pm$ 0.9 & 63.9 $\pm$ 0.6 & \textbf{65.4} $\pm$ 0.7 & \textbf{66.2} $\pm$ 1.0 \\
        $\{$intoxication$\}$ & 1.2 & 62.3 $\pm$ 0.9 & 63.4 $\pm$ 0.5 & \textbf{65.2} $\pm$ 0.6 & \textbf{66.6} $\pm$ 1.1 \\ \midrule
        $\emptyset$ & 1 & 62.6 $\pm$ 0.8 & \textbf{65.4} $\pm$ 0.5 & \textbf{68.2} $\pm$ 0.7 & \textbf{67.6} $\pm$ 1.1 \\ 
    \bottomrule
    \end{tabular}}%
}{%
  \caption{\textbf{Off-Policy Evaluation (OPE)} on the real-world medical dataset. Delphic ORL yields improvements when $z$ strongly confounds treatment decisions (large $\Gamma$). Mean and 95\% CIs over 10 runs. Best and overlapping results in bold.}\label{tab:hirid_res}%
}%
\ffigbox[\FBwidth]{%
  \includegraphics[width=0.42\textwidth]{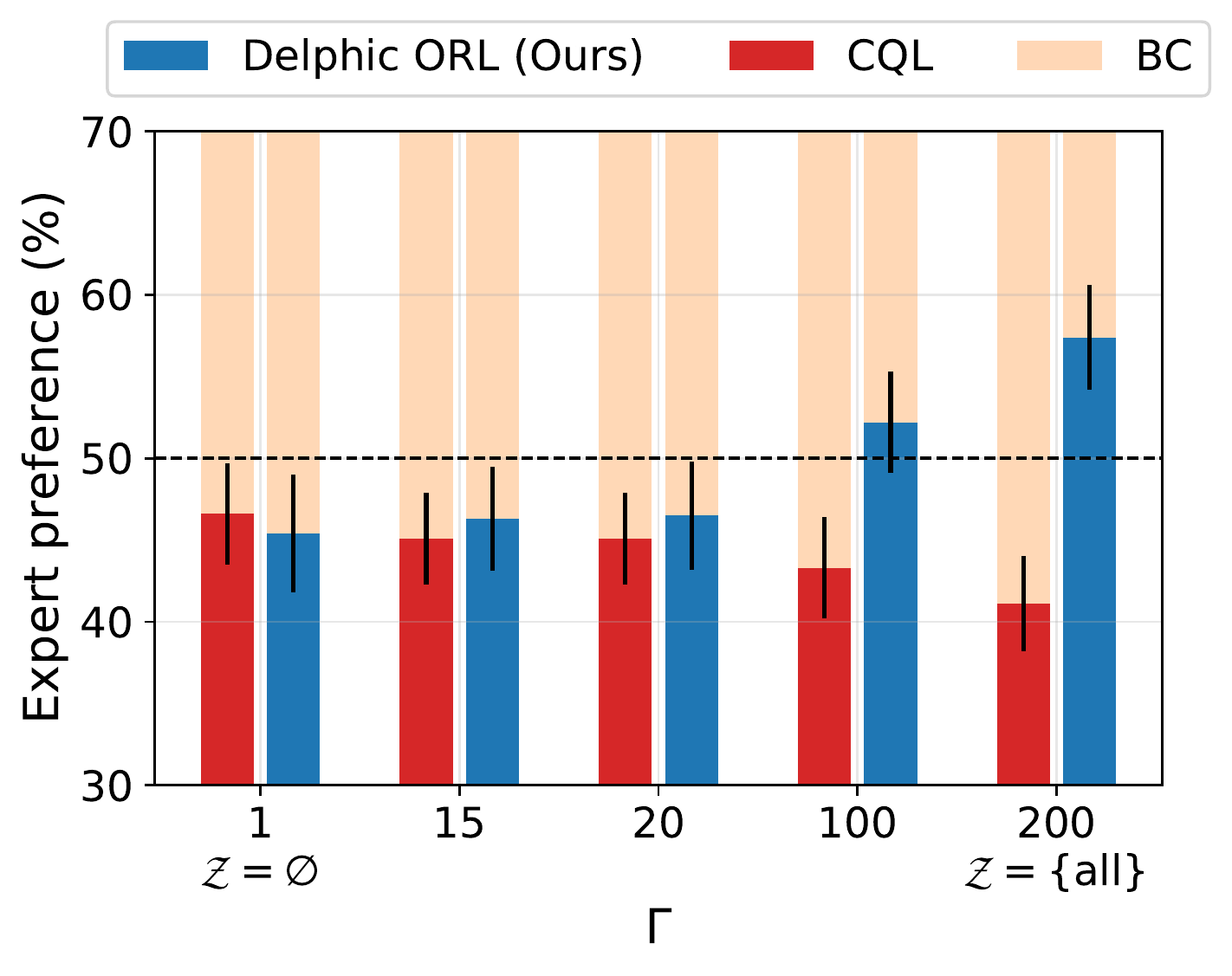}%
}{%
   \caption{\textbf{Expert Clinician Evaluation} of treatment policies, supporting the conclusion that Delphic ORL improves learning in confounded settings.}\label{fig:hirid_res_expert}%
}%

\end{floatrow}
\end{figure}

\paragraph{Confounding-Averse Policies.} We investigate the efficacy of our penalisation approach in learning policies that optimise patient outcomes in the presence of confounding bias. To evaluate the performance of these policies, we employ doubly-robust off-policy evaluation (OPE) \citep{jiang2016doubly,le2019batch}, which provides a confounding-independent estimate of treatment success by leveraging access to $z$. We refer the reader to \Cref{appendix:analysis_details} for an exhaustive overview of this evaluation method. 

\Cref{tab:hirid_res} shows our approach maintains improved performance even as the confounding level increases, while offline RL methods suffer from bias and yield suboptimal policies. As an ablation, we also studied the discrepancy of our trained policy with that in the data. Particularly, we compared the actions taken by our policy and the policy in the data and found that, unlike behaviour cloning, our policy was significantly different. This suggests our learned policy was indeed able to extrapolate from the data efficiently, identifying treatment strategies that may be more robust to confounding biases. We refer the reader to  \Cref{appendix:extra_results} for an overview of this ablation. 

\paragraph{Expert Clinician Evaluation.} Motivated by the observed success of our method, we evaluated our algorithm using expert clinicians. \Cref{fig:hirid_res_expert} shows the evaluation results of six human expert clinicians, who ranked pairs of different policies based on their observed patient outcomes. More specifically, the human experts were shown simulated patient trajectories and were asked to blindly compare the expected value of actions from either our policy or the CQL policy to those of the behaviour cloning policy. The results provide additional validation for the performance improvements of our method in confounded settings. We refer the reader to \Cref{appendix:analysis_details} for an exhaustive overview of the clinician evaluation experiment.

\section{Related Work} \label{sec:related_works}

Online-RL methods rely on environment interaction for training, limiting their applicability in many real-world domains such as healthcare \citep{gottesman2018evaluating}. This has fueled research efforts in offline methods to optimise policies through pessimism \citep{Levine2020,buckmanimportance,jin2021pessimism,xie2021bellman,ueharapessimistic,cheng2022adversarially}. Recent practical algorithmic developments in offline RL have focused on addressing statistical errors induced by epistemic and aleatoric uncertainty, in both model-based and model-free methods \citep{kumar2020conservative,kostrikov2021offline,fujimoto2021minimalist,kidambi2020morel,yu2020mopo}. 

Structural errors such as confounding bias are also pervasive in offline RL \citep{lu2018deconfounding}. Such biases cannot be captured by epistemic or aleatoric uncertainty quantification methods, as they do not depend on data quantity. Confounding bias cannot be reduced to the missing information problem in partially-observable environments either \citep{hausknecht2015deep}. History-dependent policies, for example, are equally prone to this source of error: while long-term information can recover latent environment information, it exacerbates distribution shifts between behavioural and learned policies when learning from observational data \citep{swamy2022sequence, ortega2021shaking}.

Several approaches have been proposed to address confounding bias in offline RL. Most make assumptions to estimate the confounding variables, including access to the environment \citep{lu2018deconfounding,Zhang2020DTR} or to observable back- or front-door proxy variables \citep{kumor2021sequential, wang2021provably, lu2022pessimism, shi2022off}. This allows algorithms to apply covariate adjustment methods \citep{pearl2009} to correct for confounding when modelling alternative policies (interventional probabilities and counterfactuals). Extensive work also discusses confounding bias in off-policy evaluation \citep{bennett2021proximal, bennett21ope} and bandits \citep{sen17a,tennenholtz2021bandits,chen2023unified}, but the proposed solutions remain poorly translatable to learning offline RL policies in practice, due to the aforementioned limiting assumptions.

Our work is also closely related to research on sensitivity analysis for treatment effect estimation under hidden confounding \citep{rosenbaum2002, kallus19a, jesson21a, oprescu2023b}. These works propose partial identification bounds for confounded heterogeneous treatment effect estimation or bandit decision-making problems \citep{kallus2018confounding} by assuming a bound on the dependence of the behavioural policy on hidden confounders. In this context, \citet{jesson21a} also distinguish sources of aleatoric and epistemic uncertainty from confounding biases. Other work has proposed sensitivity analysis bounds for off-policy evaluation, formulating uncertainty sets over policy returns \citep{Zhang2019, Namkoong2020}. Still, regret bounds from sensitivity analysis remain wide and often ill-adapted to high-dimensional state and action spaces or sequential decision-making problems. Our approach complements these theoretical frameworks with a practical solution to addressing confounding bias in offline RL. Finally, \citet{saengkyongam2023invariant,Tennenholtz2022} also study confounding in offline environments, but are more concerned with the complementary challenge of covariate shift -- with the latter work even assuming access to the contextual information.

\section{Conclusion}

We proposed a practical solution to address the challenge of learning from confounded data, specifically in situations where confounders are unobserved and cannot be identified. Delphic ORL captures uncertainty by modelling world models compatible with the observational distribution, achieving improved performance across both simulated and real-world confounded offline RL tasks. Our results demonstrate that Delphic ORL can learn useful policies in cases where traditional algorithms fail due to excessive confounding. Overall, we believe research into tackling hidden confounding in offline RL will lead to more reliable and effective decision-making tools in various critical fields. %

Finally, we note several limitations of our work. First, we focused our empirical evaluation on medically-motivated confounding scenarios, on the hypothesis that these should be representative of general confounded decision-making contexts. Second, the computational cost of modelling compatible worlds in Delphic ORL may be expensive for large-scale, highly confounded problems. Lastly, as with any RL algorithm, the effectiveness and safety of Delphic ORL depend on the quality and representativeness of the training data. We refer the reader to \Cref{appendix:impact} for further discussion on the limitations of our approach as well as the possible societal impact of our work.

\bibliographystyle{plainnat}
\bibliography{references}

\newpage
\appendix

\section{Additional Related Work} \label{appendix:related_works}

\subsection{Statistical Uncertainty Estimation} \label{appendix:related_works_uncertainty}

Uncertainty estimation is a crucial aspect of machine learning models, as it provides valuable insights into the reliability and confidence of model predictions and can be used to guide policy optimisation in reinforcement learning. Statistical sources of error can be estimated through aleatoric and epistemic uncertainty, which have been widely studied in the machine learning literature \citep{hullermeier2021aleatoric}. In this section, we review existing methodologies for capturing and quantifying these two types of uncertainty.

\paragraph{Aleatoric Uncertainty.} Aleatoric uncertainty, also known as data uncertainty or irreducible noise, stems from the inherent variability and randomness in the observed data \citep{hullermeier2021aleatoric}. This form of statistical uncertainty cannot be reduced even with infinite data quantities. 

The most common approach to modelling aleatoric uncertainty is to set a probability distribution over model outputs and to learn its parameters \citep{kendall2017uncertainties}. Outputs can for instance be assumed to be normally distributed with either a fixed variance (introducing a single parameter to be estimated through maximum likelihood), or a variance that depends on the input. In this latter case of heteroscedastic aleatoric uncertainty, a separate neural network branch can be trained to predict the variance \citep{kendall2017uncertainties}. %

\paragraph{Epistemic Uncertainty.}
Epistemic uncertainty arises from the lack of knowledge or ambiguity in the model parameters \citep{hullermeier2021aleatoric}, which can be reduced with additional data. Capturing epistemic uncertainty is particularly important to approximate model error in out-of-distribution scenarios \citep{jin2021pessimism}. %

Bayesian neural networks (BNNs) offer a principled approach to capturing epistemic uncertainty~\citep{neal2012bayesian}. By placing prior distributions over the model weights and using Bayesian inference, BNNs can provide posterior distributions over the weights, which represent the uncertainty in the model parameters. This uncertainty can then be propagated through the network to obtain predictive distributions that quantify epistemic uncertainty.

Bootstrap ensemble methods are another effective epistemic uncertainty estimation technique~\citep{efron1982}. These methods rely on creating multiple subsets, or bootstrapped samples, from the original dataset by randomly sampling with replacement. Each bootstrapped sample is then used to train a separate model, resulting in an ensemble of models with slightly different parameter configurations. By aggregating the predictions from these diverse models, the epistemic uncertainty can be estimated through measures such as variance or entropy. Bootstrap ensemble methods provide a practical and scalable approach to capturing model uncertainty, particularly when Bayesian methods are computationally expensive or infeasible~\citep{lakshminarayanan2017simple}. %

Monte Carlo dropout sampling~\citep{gal2016dropout} can also be used to estimate epistemic uncertainty by performing multiple forward passes with dropout enabled at test time. The distribution of predictions from these multiple samples gives an estimate of the predictive uncertainty. Finally, more recent efforts in epistemic uncertainty estimation include randomised priors~\citep{osband2018randomized}, epistemic neural networks \citep{osband2021} and deep ensembles trained with Stein variational gradient descent \citep{liu2016stein,dangelo2021}.

\subsection{Nonidentifiable Confounding Bias} \label{appendix:causal_related_work}

\paragraph{Comparison to Sensitivity Analysis.}
In the causal inference literature, sensitivity analysis studies the robustness of treatment effect estimation to hidden confounding. This framework assumes a bound on the ratio between treatment propensities between any two confounder values \citep{rosenbaum2002} or on the ratio between treatment propensities when accounting for and marginalising confounders \citep{kallus19a,jesson2020,oprescu2023b}. %

In contrast, the main assumptions in our delphic uncertainty estimate determine which `world models' compatible with the observational data are considered to construct uncertainty sets over a given outcome model (Conditional Average Treatment Effect or, in the sequential setting, Q-value function). In particular, we consider a set of possible $\Zspace$ and prior distributions $p(z)$, and specify a model architecture for the dependence of the behavioural policy and transition, reward or value function on $z$ (which is then trained to fit the marginalised observational trajectory distribution).

Importantly, sensitivity analysis approaches require domain expertise to set maximum propensity-ratio parameter $\Gamma$ \citep{kallus19a}, from which uncertainty sets over the modelled outcome are derived. Delphic ORL does come with its own set of hyperparameters (number of world models considered in $u_d$, pessimism hyperparameter $\lambda$), which can be determined through practical, quantitative means with arguably less domain expertise.

\paragraph{Time-Varying Confounders.} The Contextual Markov Decision Process \citep{Hallak2015} and associated problem described in \Cref{sec: preliminaries} describes confounders as sampled from a context distribution $\nu(z)$ and fixed over the course of an episode. We note that this framework does not exclude the existence of time-varying confounders. Consider the Markov Decision Process with Unmeasured Confounding (MDPUC) \citep{zhang2016markov}, in which a new i.i.d. hidden confounder variable $z_t$ affects the transition at each timestep $t$. This framework can be framed as a CMDP where the overall episode context $z=\{z_1, \ldots, z_H\}$ includes all confounder variables. Although we do not focus on this specific framework in our experimental setting, this would form an interesting avenue for further work. An important distinction between MDPUC and Partially-Observable Markov Decision Processes (POMDPs) is the assumption that confounder variables are sampled i.i.d. at each timestep. While POMDPs can therefore be viewed as the generalisation of this decision-making setup, note that confounding biases are only induced in this setup if the behavioural policy has access to some missing information about the state variable.

\subsection{Broader Impact, Limitations and Future Work} \label{appendix:impact}

Addressing hidden confounding in offline reinforcement learning has the potential to significantly impact the development and deployment of reinforcement learning systems in real-world applications. By improving the validity of causal conclusions drawn from data, Delphic ORL can improve the effectiveness and safety of RL-based decision-making in critical fields \citep{thomas2017predictive,gottesman2018evaluating,singla2021reinforcement}.

While our results demonstrate the efficacy of Delphic ORL in learning useful policies in the presence of confounding, it is important to acknowledge the limitations and potential unintended consequences associated with RL algorithms, especially in high-stakes applications such as healthcare. Collaboration with domain experts is crucial to ensure thorough evaluation of RL algorithms \citep{gottesman2018evaluating}. In clinical settings, predictive or recommendation models derived from Delphic ORL should not be solely relied upon, and mitigation strategies must be implemented to minimise negative consequences during deployment.

An important consideration in the application of Delphic ORL is the trade-off between confounding bias and estimation variance in Q-function estimation, as noted in other work addressing confounding bias \citep{wang2019blessings}. This emphasizes the significance of large, high-quality training datasets to leverage the benefits of Delphic ORL and ensure sufficient predictive power.

As our experiments primarily focused on medically-motivated confounding scenarios, future work should investigate the applicability and generalisation of Delphic ORL to other domains. Although our framework does not in theory exclude dynamic environments where confounding factors change over time (see \Cref{appendix:causal_related_work}), an empirical study of the behaviour of delphic uncertainty estimates and pessimism penalties may reveal new challenges in this context.

Finally, the question of how to best approximate the set of compatible worlds $\W$ in \Cref{def: compatible world models} remains open. In \Cref{sec:method} and \Cref{appendix:implementation}, we detail our approach which efficiently captures variability across counterfactual value-function, but further theoretical or practical work on how best to model $\W$ would likely improve the calibration of delphic uncertainty estimates. Better approximation algorithms may also improve the efficiency, scalability, and modelling power of our method for very high-dimensional, highly confounded problems -- although our real-world data analysis forms a promising first proof-of-concept.

\section{Theoretical Details} \label{appendix:proofs}

\subsection{Proof of \Cref{thm:var_decomp}}

We start by considering the decomposition of variance in $Q_\theta^\pi$ caused by random variable $\theta$. In the following, we drop superscript $\pi$ for clarity.

First, we decompose $\Var(Q_\theta \mid \theta)$:
\begin{align}
    \Var(Q_\theta \mid \theta) &= \Expt[Q_\theta^2 | \theta] - \Expt[Q_\theta| \theta]^2  \nonumber\\
    \Expt_\theta[\Var(Q_\theta \mid \theta)] &= \Expt_\theta[\Expt[Q_\theta^2 | \theta]] - \Expt_\theta[\Expt[Q_\theta| \theta]^2]  \nonumber \\
    & =\Expt[Q_\theta^2] - \Expt_\theta[\Expt[Q_\theta| \theta]^2]  \label{eq:vary}
\end{align}
where the last line results from the law of iterated expectations: $\Expt_B [ \Expt [A|B] ] = \Expt[A]$ for two random variables $A,B$.

Next, we study $\Var(\Expt[Q_\theta \mid \theta])$:
\begin{align}
    \Var_\theta(\Expt[Q_\theta \mid \theta]) &= \Expt_\theta[\Expt[Q_\theta \mid \theta]^2] - \Expt_\theta[\Expt[Q_\theta \mid \theta]]^2 \nonumber \\
     &= \Expt_\theta[\Expt[Q_\theta \mid \theta]^2]  - \Expt[Q_\theta]^2  \label{eq:varexpty}
\end{align}
again using iterated expectations.

Summing equations \ref{eq:vary} and \ref{eq:varexpty}, we obtain:
\begin{align}
    \Expt_\theta[\Var(Q_\theta \mid \theta)]  + \Var_\theta(\Expt[Q_\theta \mid \theta]) &=\Expt[Q_\theta^2]  - \Expt[Q_\theta]^2 \nonumber \\
    &= \Var(Q_\theta) \label{eq:total_var_basic}
\end{align}

This result is known as the law of total variance \citep{weiss2006course}, which can be interpreted as a decomposition of epistemic and aleatoric uncertainty \citep{kendall2017uncertainties}.

We can rewrite the above result within a given world model $w$, denoting $\theta$ as $\theta_w$. Now conditioning on the world model $w$, we have:
\begin{equation} \label{eq:total_var_theta_w}
\Var(Q_{\theta_w} \mid w) = \Expt_{\theta_w}[\Var(Q_{\theta_w} \mid \theta_w, w) | w]  + \Var_{\theta_w}(\Expt[Q_{\theta_w} \mid \theta_w, w] | w)
\end{equation}

We also write equation \ref{eq:total_var_basic} such that the conditioning random variable is now $w$, which induces variation in $Q_{\theta_w}$ if we consider a counterfactual trajectory distribution. Combined with \Cref{eq:total_var_theta_w}, we obtain:
\begin{align}
     \Var(Q_{\theta_w}) &= \Expt_w[\Var(Q_{\theta_w}|w)] 
     + \Var_w \left(\Expt[Q_{\theta_w} \mid w ]\right) \nonumber \\
     &= \Expt_w \brk[s]2{
    \Expt_{\theta_w}[\Var(Q_{\theta_w} \mid \theta_w, w) | w]  + \Var_{\theta_w}(\Expt[Q_{\theta_w} \mid \theta_w, w] | w)
     } \nonumber \\
     & \hspace{4cm} + \Var_w \brk2{
     \Expt_{\theta_w} \left[ \Expt[Q_{\theta_w} \mid \theta_w, w ] | w \right]
     } 
\end{align}
using iterated expectations. This concludes the proof of \Cref{thm:var_decomp}.

Note that if we assume $Q_{\theta_w}$ has a Gaussian distribution for fixed $\{\theta_w, w\}$, parameterised as $\mathcal{N} (\mu_{\theta_w}, \sigma^2_{\theta_w})$, we have $\Var (Q_{\theta_w}\mid \theta_w, w) = \sigma_{\theta_w}^2$ and $\Expt [Q_{\theta_w}\mid \theta_w, w] = \mu_{\theta_w}$.
We obtain results in \Cref{eq:total_var} by expanding the first term in the variance decomposition, $\Expt_{\theta_w}[\sigma_{\theta_w}^2 |w]$, as follows:
\begin{align*}
\Expt_{\theta_w}[\sigma_{\theta_w}^2 |w] &= \Expt_{\theta_w}[\sigma_{\theta_w}^2 |w] -\Expt_{\theta_w}[\sigma_{\theta_w} |w]^2 + \Expt_{\theta_w}[\sigma_{\theta_w} |w]^2 \\
&= \Var_{\theta_w}(\sigma_{\theta_w} |w) + \Expt_{\theta_w}[\sigma_{\theta_w} |w]^2.   
\end{align*}

\subsection{Asymptotic Interpretation of \Cref{thm:var_decomp}} \label{appendix:asymptotic} 
 We consider three extreme cases of \Cref{thm:var_decomp} to clarify its decomposition. First, we consider the limit of infinite-data with no confounding (e.g., no dependence on $z$). In this case, $\theta_w$ and $w$ converge to a single ground-truth. Any remaining statistical error will come from the intrinsic environment stochasticity or the behavioural policy, and therefore has an aleatoric nature. Indeed, only the first term in \Cref{thm:var_decomp} would remain.

Next, consider the setting in which the value is a deterministic mapping of states, with only one compatible world model. Learning from finite data quantities leads to statistical error in optimising the parameters $\theta_w$, and is known as epistemic uncertainty. Indeed, deterministic environments with only one compatible world model will reduce \Cref{thm:var_decomp} to the second term.

Finally, we consider the case of infinite data in a deterministic setting. In this case, multiple compatible world models may exist which induce the same observational distribution (as demonstrated in \Cref{sec:sources of error}). The source of error remaining is delphic uncertainty, and arises if multiple models assign high likelihood to the observational data, but return different estimates of the value. In this paper we propose to estimate this final form of uncertainty by learning an ensemble of compatible world models, in a similar fashion to the bootstrap method for quantifying epistemic uncertainty.

\section{Implementation Details} \label{appendix:implementation}

\subsection{Statistical \& Delphic Sources of Uncertainty}

\paragraph{World Model Training.} We implement world models as variational models for estimating the confounder distribution, jointly with a model for the behaviour policy $\behav$ and for the action-value function $Q^{\behav}$, both dependent on a $z$ sampled from the posterior. As the environments we consider have discrete action spaces, we learn the behaviour policy by minimising its cross-entropy on the training data, as in behaviour cloning. Training is carried out for 50 epochs or until loss on the validation subset (10\% of training data) increases for more than 5 consecutive epochs. Within a world model $w$, hyperparameters $\{ \alpha, \beta\}$ can be tuned based on prediction performance on the validation set. %

Model $Q^{\behav}$ corresponds to an on-policy action-value function approximation. We compute targets through Monte Carlo updates (for the sepsis environment with sparse episodic rewards) or Temporal Difference learning (for the real-world ICU dataset) based on samples from the observational training data with a discount factor of $\gamma=0.99$. The Q-function is trained as a classifier over 200 quantiles.

Between world models $w$, the confounder space dimensionality is randomly varied over $|\Zspace| = \{1, 2, 4, 8, 16\}$, and the prior for $p(z) = \mathcal{N}(z; 0, \Sigma^2)$ is randomly varied through the variance for each $z$-dimension, $\Sigma^2_{ii} = \{1.0, 0.1, 0.01 \}$. For the sepsis simulation, the encoder architecture for the confounder distribution $\nu(z|\tau)$) consists of a multi-layer perceptron with hidden layer dimensions $(128, 64, 32)$ and ReLU activation before the final layer mapping to dimension $|\Zspace|$. For the real dataset, the encoder architecture  is implemented as a transformer \citep{Vaswani2017} with 2 layers, 4 heads, and embedding dimension 32, considering a maximum history length of 10 tokens. The behavioural policy $\behav(a|s,z)$ and action-value function $Q^{\behav}(s,a,z)$ are both implemented as multilayer perceptrons with hidden layer dimensions $(32, 64, 128)$ and ReLU activation.

\paragraph{Uncertainty Estimates.}
Additional inductive biases can be incorporated to capture epistemic and aleatoric uncertainty within a single world model $w$, as these relate to {statistical} sources of uncertainty. Following prior work \citep{kendall2017uncertainties, yu2020mopo}, we capture aleatoric uncertainty by modelling a normal probability distribution over outputs $(\behav, Q^{\behav})$. We then measure epistemic uncertainty within each world model $w$ by training on different data bootstraps, returning an ensemble of parameters $\{\theta_w^1, \theta_w^2, \ldots\}$ for each $w$. %

Recalling \Cref{eq:total_var}, the delphic uncertainty term $\Var_{w} ( \Expt_{\theta} [\mu_{\theta_w} ] )$ is estimated by measuring the variance between  predictions $\mu_{\theta_w}$ (averaged over model parameters $\theta_w$), across across multiple generative models $w$. Epistemic uncertainty can be estimated as the variance of outputs over different model parameters $\theta_w$, averaged across worlds $w \in \mathcal{Z}$.
Finally, aleatoric uncertainty is measured through the fitted probability distribution over model outputs $Q^{\behav}_{\theta_w}$, averaged over all $\theta_w$ in a given world, and over all worlds $w \in \W$.

The number of world models $W$ was varied between 5 and 20 for both datasets and was chosen as the smallest number converging to an average delphic uncertainty comparable to the largest $W$. An ablation of delphic uncertainty as a function of the number of world models is given in \Cref{appendix:extra_results}. This resulted in 10 and 15 world models for the sepsis and real-world datasets respectively. Finally, each world model was trained over 5 different data bootstraps to estimate epistemic uncertainty. Overall, compared to sensitivity analysis where parameter $\Gamma$ needs to be fixed through domain expertise \citep{oprescu2023b}, we found delphic uncertainty to be less dependent on expert input in determining model parameters.

\paragraph{Counterfactual Estimates.} Note that while our approach changes the policy term in $P_{\behav}$ to obtain counterfactual estimates, other factors in the world model (e.g. $\nu_w$, $Q^{\behav}_{w}$) could be varied to obtain general counterfactual predictions in this world model. 
As an example, we also found promising results by measuring delphic uncertainty through variation across $w$ over the following counterfactual quantity: $\Expt_{(s,a) \in \dataset}\Expt_{z \sim \textcolor{TabBlue}{p_{w}(z)}} \Expt_{\theta_w}[Q^{\behav}_{\theta_w}(s,a,z)]$, where $z$ is sampled from the model prior $p_{w}(z)$
instead of the learned posterior $\nu_{\theta_w}(z|\tau)$. In this case, the resulting delphic uncertainty estimate, capturing variation over the counterfactual quantity across world models, becomes independent of a given policy -- and dependent on the new quantity introduced (in the previous example, on prior $p_{w}(z)$).

\subsection{Delphic Offline Reinforcement Learning}

\begin{algorithm}[t!]
        \caption{Delphic Offline Reinforcement Learning: \textcolor{TabBlue}{Bellman Penalty} in Offline Q-Learning Algorithm.} \label{alg:algo_qtarget} 
        \SetKw{Input}{Input:}
        \Input{Observational dataset $\dataset$, Model-free Offline RL algorithm (e.g. CQL \citep{kumar2020conservative}), Penalty hyperparameter $\lambda$.} \\
        Learn a set of compatible world models $\{\Zspace_w, \nu_w, \rho_{0,w}, P_{r,w}, T_w, \pi_{b,w}\}_{w\in \W}$ that all factorise to $P_{\behav}(\tau)$. \\ 
        Obtain counterfactual predictions $Q_w ^{\pi}$ for each $w \in \W$. \\
        Define local delphic uncertainty: $u_d^{\pi} (s,a) = \Var_w ( Q_w ^{\pi} (s,a) )$. \\
        Initialise Q-function parameters $\phi$. \\
        \For{each iteration}{
        Sample $(s,a,r,s') \sim \dataset$. \\
        Compute penalised Bellman target: $Q_{target}' = r + \gamma \max_{a' \in \Aspace} Q_{\phi}(s', a') \textcolor{TabBlue}{- \lambda u_d^{\pi} (s, a)}$, where $\pi(a|s) = \text{argmax}_a Q_{\phi}(s,a)$. \label{algoline:u_penalty} \\ %
        Perform gradient descent  w.r.t. $\phi$ on $\left[ Q_{\phi}(s,a) - Q_{target}'(s,a) \right]^2 + \mathcal{R}_{offline}(\phi) $, where regularisation term $\mathcal{R}_{offline}$ depends on the choice of offline learning algorithm. \\
        }
\end{algorithm}

We detail our learning procedure in \Cref{alg:algo_qtarget}. As our base offline RL algorithm is CQL \citep{kumar2020conservative},  our regularisation term $\mathcal{R}_{offline}(\phi)$ is the CQL penalty: $\mathcal{R}_{offline}(\phi) = \alpha \left[ \log \sum_{\tilde{a}\in \Aspace} \exp Q_{\phi}(s,\tilde{a}) - Q_{\phi}(s,a) \right]$. We base our algorithm on an existing implementation for CQL \citep{d3rlpy}, which includes additional training details for stability, such as target networks, double Q-networks and delayed updates \citep{fujimoto2018td3}. For architecture details, see the baseline implementation of CQL in \Cref{appendix:baselines}. As for all baseline algorithms, we train for $100$ epochs, using 500 (sepsis dataset) or $10^4$ (ICU dataset) timesteps per epoch. In practice, the policy $\pi$ considered for uncertainty estimation and the target network are updated every 8000 timesteps, to improve stability in training. 

Note that an actor-critic variant of \Cref{alg:algo_qtarget} is also feasible, setting $\pi$ in $u_d^{\pi}$ to be the actor policy, as well as other offline learning paradigms in $\mathcal{R}_{offline}(\phi)$, such as BC regularisation \citep{fujimoto2021minimalist}.

\paragraph{Alternative Forms of Pessimism.} Following the discussion on alternative forms of pessimism in \Cref{sec:pessimism}, we propose practical alternatives to the Delphic ORL penalty in \Cref{algoline:u_penalty} of \Cref{alg:algo_qtarget}, which substracts a factor of $u_d$ from the Q-function Bellman target based. In the following, note that $u_d$ can also be independent of $\pi$ if varying over different factors in $P_{\behav}$ as detailed above. %

\begin{itemize}
    \item \textbf{Delphic ORL via Uncertainty Threshold:} One approach, inspired by Batch Constrained Q-Learning \citep{fujimoto2019off}, is to constrain value function updates to only consider actions falling below a certainty uncertainty threshold. For a tuple $(s,a,r,s')$, the Q-function Bellman target  can be computed as: $Q_{target}' = r + \gamma \max_{a':  \textcolor{TabBlue}{u_d^{\pi}(s',a') < \lambda}} Q_{\phi}(s', a')$, where $\lambda$ is a threshold controlling the maximum delphic uncertainty accepted for a given action choice. 

    \item \textbf{Model-Based Delphic ORL:} In model-based methods, a penalty proportional to the uncertainty $u_d(s,a)$ can be substracted from the reward function $r(s,a)$, as in \citet{yu2020mopo}. The effective reward function becomes: $\tilde{r}(s,a) = r(s,a) \textcolor{TabBlue}{- \lambda u_d (s,a)}$.

    \item \textbf{Delphic ORL via Weighting:} The uncertainty measure can also be used to weight samples in the objective function, prioritising unconfounded states and actions during training:     \begin{equation*}
    \Expt_{(s,a,r) \sim \dataset} \left[ \textcolor{TabBlue}{\frac{\lambda}{u_d(s,a)}} \mathcal{L}(s,a,r) \right]
    \end{equation*}
    where $\mathcal{L}$ can be the Q-function Bellman update or the supervised learning objective for behaviour cloning.
\end{itemize}
We compare the performance of different implementations of pessimism on the simulated sepsis environment in \Cref{appendix:extra_results}.

\paragraph{Hyperparameter Tuning.} There is no natural validation criterion in Offline RL, and the best approach to choose hyperparameters in this context remains an open question \citep{Levine2020}. In practice, we run our algorithm for 4 different values of $\lambda \in \{10^{-3}, 10^{-2}, 10^{-1}, 1\}$ and choose the final policy giving the best off-policy evaluation performance on the validation set (using the Fitted Q-Evaluation implementation available in the codebase, \citet{le2019batch}). As noted in related works, expert input may be useful at this stage to also determine how strong a penalty again potential hidden confounding would be desirable or how much confounding could be expected \citep{rosenbaum2002}. Other hyperparameters specific to offline RL algorithms are tuned in the same way and are given in the following section. %

\subsection{Baseline Methods \& Training Details} \label{appendix:baselines}

All reinforcement learning algorithms and baselines are implemented based on the open access \texttt{d3rlpy} library \citep{d3rlpy}. The discount factor used is $\gamma=0.99$, and state and actions are normalised to mean 0 and variance 1 \citep{fujimoto2021minimalist} for all algorithms. Training is carried out on NVIDIA RTX2080Ti GPUs on our local cluster, using the Adam optimiser with default learning rate and a batch size of 32. Models are trained for 100 epochs with 500 (sepsis dataset) or $10^4$ (ICU dataset) timesteps per epoch. Model-specific hyperparameters are tuned as in Delphic ORL.

\paragraph{Behaviour Cloning (BC).} Behaviour cloning \citep{Ross2011} is a supervised learning model of the behaviour policy, mapping states to actions observed in the dataset. After considering the following architectures: multi-layer perceptron (MLP), Long Short Term Memory (LSTM) network \citep{hochreiter1997long}, Gated Recurrent Unit (GRU) \citep{cho2014learning} and Transformer \citep{Vaswani2017}, GRU was found to give the best validation performance on both the simulated and real datasets. Implementation details for the GRU BC models include two hidden layers of dimension $(64, 32)$ and ReLU activation. The last layer is passed through a softmax layer to produce action probability outputs, and the model is trained by minimising action cross-entropy over the observational dataset, with L2 regularisation of weight 0.01.

\paragraph{Conservative Q-Learning (CQL).} Discrete CQL \citep{Kumar2019} is implemented with a penalty hyperparameter $\alpha$ of 1.0 (sepsis environment) and 0.5 (ICU dataset), tuned over the following values: $\{0.1, 0.5, 1.0, 2.0, 5.0 \}$. The Q-function is implemented as a distributional model with a standard MLP architecture (two linear layers with 256 hidden units) and 200 quantile regression outputs.

\paragraph{Batch Constrained Q-Learning (BCQ).} Discrete BCQ \citep{fujimoto2019off} is implemented with a threshold for action flexibility set to 0.5 for both environments, tuned over the following values: $\{0.1, 0.3, 0.5, 1.0, 2.0, 5.0 \}$. %
The Q-function is implemented as a distributional model with a standard MLP architecture (two linear layers with 256 hidden units) and 200 quantile regression outputs.

\section{Experimental Details} \label{appendix:expdetails}

\subsection{Decision-Making Environments} \label{appendix:data}

\paragraph{Sepsis Environment.} Introduced by \citet{Oberst2019}, this environment simulates the trajectory of patients in the intensive care. Based on the authors' publicly available code\footnote{\url{https://github.com/clinicalml/gumbel-max-scm}}, our state space $\Sspace$ consists of
4-dimensional observation vectors (measures for heart rate, systolic blood pressure, oxgenation and blood glucose levels) which we normalise to mean and variance $(0,1)$. The discrete action space $\Aspace$ comprises the combination of three binary treatments (antibiotic, vasopressor or ventilation administration) for a total dimension of 8. An unobserved binary variable $z$ encodes the diabetic status of patients, with 20\% of trajectories having a positive status. The agent obtains a reward of $+1$ if the patient reaches a healthy state (and is thus ready for discharge) and a negative reward of $-1$ if the patient reaches a death state.

The observational dataset $\dataset$ is generated by rolling out the optimal (diabetes-aware) policy in the environment for 10,000 environment interaction steps, taking a random action with probability $\epsilon = 0.1$ to ensure sufficient state-action coverage for offline learning. The maximum episode length is set to 20 timesteps. The resulting dataset has a confounding strength of $\Gamma=100$.

Environment stochasticity can be varied by changing the variance around the originally deterministic reward obtained at the end of a trajectory, between $\sigma^2_r=0$ as in the original environment and $\sigma^2_r = 0.4$. Datasets of varying confounding strength $\Gamma \in [1, 100]$ are obtained by setting the behaviour policy for $z=1$ as a weighted average of the policies for different $z$ values: $(1-p) \behav(z=0) + p\behav(z=1)$, where $p$ depends on $\Gamma$ and $\epsilon$. Environment transition and reward functions and their dependence on $z$ are kept fixed. Finally, we vary the dimension of the confounder space $\Zspace$ by introducing more binary indicators with the same effect on the transition dynamics as the diabetes indicator. %

\paragraph{Electronic Health Records Dataset.} Our real-world data experiment is based on the publicly available HiRID dataset \citep{Hyland2020}. This dataset counts over 33 thousand patient admissions at an intensive care unit in Bern University Hospital, Switzerland \citep{Hyland2020} and can be pre-processed using open access code from the HiRID benchmark \citep{Yeche2021}. Patient stays were downsampled to hourly measurements and truncated to a maximum length of 20 hours and default training, validation and test sets were used.

We consider the task of optimising fluid and vasopressor administration ($\Aspace$ is the combination of two binary choices). The reward function is designed to penalise circulatory failure events ($r=-1$ for all timepoints in the duration of the event) and to reward timepoints where the patient is not in such a critical state ($r=1$, and $r=2$ in the timepoint following recovery from circulatory failure). %
Circulatory failure events for each patient are labelled following internationally accepted criteria \citep{Yeche2021}. This short-term reward function is dense, unlike previous RL work on optimising intravenous fluid and vasopressor administration \citep{Raghu2017}, making off-policy evaluation more reliable \citep{gottesman2018evaluating}. 

The state space $\Sspace$ consists of all variables in the electronic health records which are not considered treatment for the organ system considered, based on the variable categorisation released with the dataset \citep{Hyland2020}. This results in a state space dimensionality of 203. The list of variables excluded for each task in given in Table~\ref{tab:hirid_details}. At each timepoint within a patient stay, we also compute the Sequential Organ Failure Assessment (SOFA) score \citep{vincent1996sofa} which is used to quantify the severity of a patient's illness in the intensive care unit. A higher score indicates greater severity of illness. %

\begin{table}[]
    \centering
    \caption{\textbf{Offline reinforcement learning task on real-world medical dataset.}}
    \label{tab:hirid_details}

\begin{tabular}{lll}
\toprule
Task                                       & Circulatory treatment      &                                                \\
\midrule
Action space $\Aspace = \{0,1\}^2$           & Fluids                     & Vasopressors                                   \\
\midrule
Organ failure avoided by $R$               & \multicolumn{2}{l}{Circulatory failure}                                     \\
\midrule
State space $\Sspace$ (selected variables, & Heart rate                 & Respiratory rate                               \\
$|\Sspace| = 204$)                         & Body temperature           & Urinary output                                 \\
                                           & Blood pressure             & GCS score                                      \\
                                           & Cardiac output             & Central venous pressure                        \\
                                           & Oxygen saturation          & Base excess                                    \\
                                           & Lactate                    & Arterial pH                                    \\
                                           & PaO2                       & Creatinine                                     \\
                                           & Serum sodium               & Serum potassium                                \\
                                           & Haemoglobin                & Glucose                                        \\
                                           & Other lab values           & Ventilator settings \\
                                           & Antibiotics                & Steroids                                       \\
                                           & Diuretics                  & Insulin                                        \\
                                           & Cerebrospinal fluid drain  & Anticoagulants                                 \\
                                           & \multicolumn{2}{c}{$\ldots$}                                       \\
\midrule
Other treatment variables excluded         & Blood product infusions    &  Vasodilators                  \\
                                           & Cristalloid infusion       & Antiarrhymic agents                            \\
                                           & Colloid infusion           &  Antihypertensive agents                                          \\
\midrule
Confounder variables $z$                   & Age                        & Cardiovascular diagnosis                       \\
                                           & Weight                     & Pulmonary diagnosis                            \\
                                           & Gastrointestinal diagnosis & Orthopaedic diagnosis                           \\
                                           & Neurological diagnosis     & Metabolic/endocrine diagnosis                  \\
                                           & Hematology diagnosis       & Trauma diagnosis                               \\
                                           & Sedation           & Intoxication                                   \\
                                           & Emergency status           & Surgical status                               \\
       \bottomrule
    \end{tabular}%
\end{table}

Selected confounders are obtained by excluding some state dimensions from the observational dataset (up to $|\Zspace| = 14$). These variables do not constitute the \textit{entire} confounder space, as much exogenous, unrecorded information affects patient evolution and is taken into account in medical treatment decisions \citep{Yang2018}. We ignore this in our analysis as we cannot evaluate with respect to this missing information, but we note that this is precisely the motivation behind our work.

The confounding strength $\Gamma$ for each confounding space $\Zspace$ considered was estimated as follows. Each point in the training dataset was binned into a $(s,a,z)$ category, depending on its discrete action and context values $(a,z)$ and on its SOFA score as a summary variable for $s$. We discretise the SOFA score into 5 quantiles. Finally, we compute the mean policy value for each $(s,a,z)$ bin through $\behav(a|s,z) = {P(a,s,z)}/{P(s,z)}$, and we take $\Gamma$ as the ratio $\max_{z, z'} [{\behav(a|s,z)}/{\behav(a|s,z')}]$.

\subsection{Analysis Details} \label{appendix:analysis_details}

In this section, we provide additional details pertaining to the analysis of our experimental results. All results reported in this work include 95\% confidence intervals around the mean, computed over ten training runs unless otherwise stated. Environment returns and off-policy evaluation results are normalised on a scale of 0 to 100. \Cref{fig:u_vs_NZ} was obtained by varying the dimension on the x-axis, while keeping the other variables fixed to $N=864$ trajectories, confounding strength $\Gamma=15$ and reward function variance $\sigma_r^2=0.0$. To generate \Cref{fig:confounding_ICU}, patients with the relevant confounder ($z$) value were binned by disease severity and the probability of vasopressor prescription (top) and the overall density (bottom) in each group were computed. \Cref{fig:u_increase_vs_z} was then obtained by computing the relative increase in delphic uncertainty when including the relevant $z$-dimension to the hidden context space $\Zspace$ (in other words, removing this dimension from the visible state space).

\paragraph{Off-Policy Evaluation (OPE).} Doubly robust methods trade off bias of an approximate reward model and of weighted methods with the high variance of importance sampling approaches \citep{jiang2016doubly}. Assuming $z$ is accessible for each trajectory at \textit{evaluation} time to overcome confounding, doubly-robust off-policy evaluation estimates the value of policy $\tilde{\pi}$ as follows: 
\begin{equation}\label{eq:wdr_ope}
V_{DR}(\tilde{\pi}) = \Expt_{(s,a,r,z) \in \dataset} \left[  \frac{\tilde{\pi}(a|s)}{\hat{\pi}_{b}(a|s,z)} \{r - Q(s,a,z) \}+ Q(s,\tilde{\pi}(s),z)\right], \end{equation}
where $\hat{\pi}_{b}$ is a model for the behavioural policy and $Q$ for expected returns under $\tilde{\pi}$, learned on the dataset with observable $z$. %

Fitted Q-Evaluation is an established value estimation method \citep{le2019batch}. The algorithm iteratively applies the Bellman equation to compute bootstrapping targets for Q-function updates: $Q_{k+1} \leftarrow \argmin_{Q} \Expt_{(s,a,r,z) \in \dataset} \left[ \{ r - Q(s,a,z) + \gamma Q_k(s', \tilde{\pi}(s'), z) \}^2 \right]$ which can be solved as a supervised learning problem. This results in a learned Q-value for the evaluated policy $Q^{\tilde{\pi}} (s,a,z)$ which can be used in the weighted doubly-robust estimate in \Cref{eq:wdr_ope} to provide return estimates in \Cref{tab:hirid_res}. 

Both the Q-function and the behaviour policy in \Cref{eq:wdr_ope} are parametrised as a fully-connected neural network dimension with 3 layers of hidden dimension (64, 32, 16) and ReLU activation. The former is trained by minimising the mean squared error with the Q-function update above, the latter by minimising the cross-entropy with respect to action choices in $\dataset$. %

\paragraph{Human Policy Evaluation.} Off-policy evaluation has limitations, being itself prone to its own set of statistical errors and data-related concerns \citep{gottesman2018evaluating}. We aim to confirm conclusions drawn over OPE returns through a human expert evaluation of treatment policies. %

\begin{figure}
    \centering
    \includegraphics[trim={0 3cm 0 3cm},clip, width=0.8\textwidth]{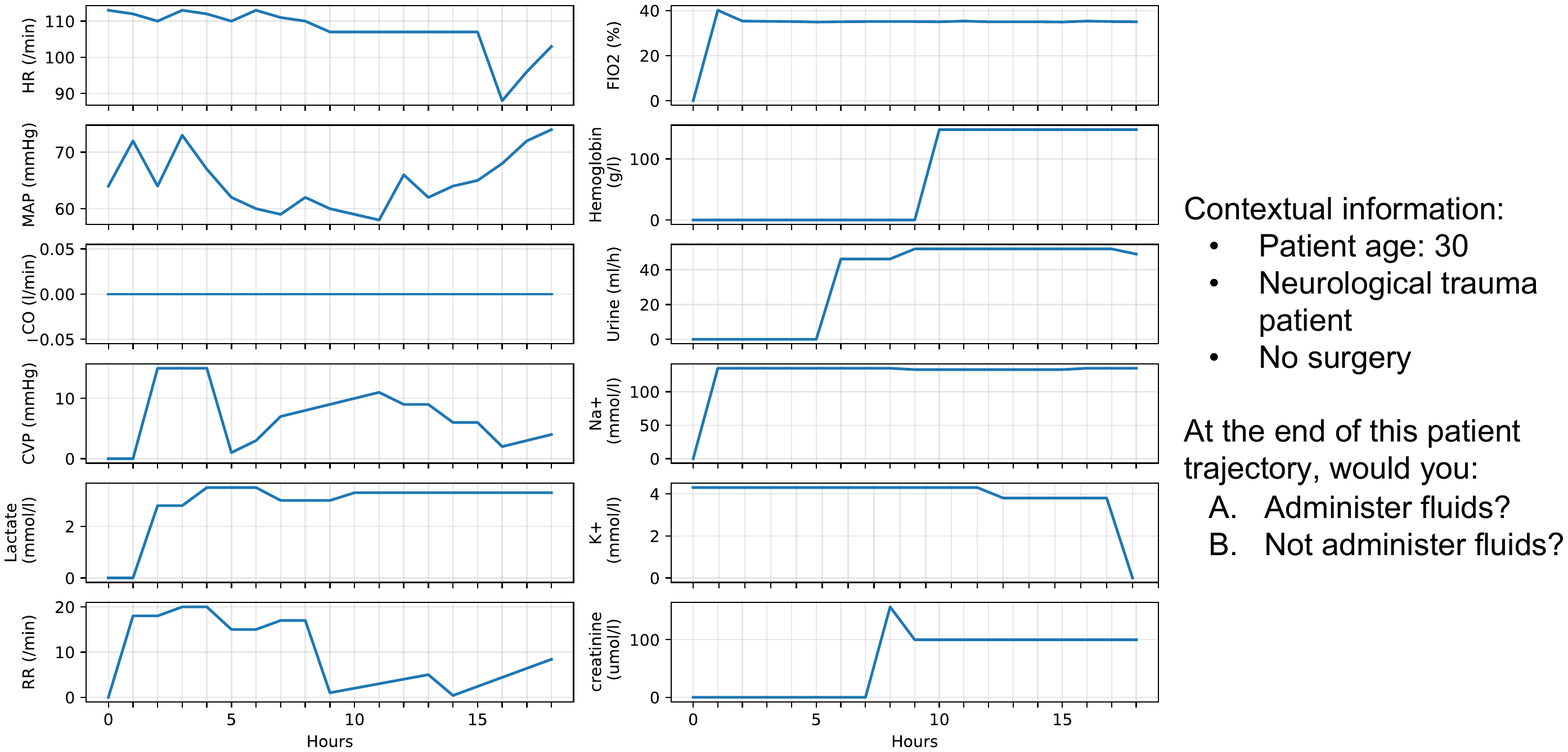}
    \caption{\textbf{Illustration of action ranking by medical experts.} Synthetic patient trajectories and a varying degree of contextual information (varying $|\Zspace|$ and $\Gamma$) are given to clinicians, who must rank the treatment options in terms of expected patient outcomes.}
    \label{fig:human_eval}
\end{figure}

Synthetic patient trajectories are first generated by randomly sampling from the ICU dataset along each state dimension, with varying amounts of contextual information as detailed in \Cref{tab:expert_eval_gamma}. Action choices at the end of the trajectories are computed for the Delphic ORL, CQL and BC policies, trained on the observational dataset with the same degree of confounding. Trajectories are selected if they induced a disagreement between these methods, to shed light on potential improvements or harmful behaviour learned by the offline RL models. Trajectories are then simplified into 12 critical variables (as shown in \Cref{fig:human_eval}), and shown to physicians, who are asked to rank two treatment options in terms of expected patient outcomes. Unknown to the physicians, and in a random order, one of the options was predicted by the Delphic ORL or CQL policy, and the other by the BC baseline. Overall, we consulted six clinicians with different degrees of expertise in intensive care (from junior assistant doctors to department heads) from Switzerland and the United Kingdom, collecting their treatment preferences over 45 such trajectories. %

\begin{table}[h]
    \centering
    \resizebox{0.8\textwidth}{!}{
    \begin{tabularx}{\linewidth}{*{6}{X}}
    \toprule
         $\mathbf{\Gamma}$ & \textbf{1} & \textbf{15} & \textbf{20} & \textbf{100} & \textbf{200} \\ \midrule
         $|\Zspace|$ & 0 & 10 & 11 & 13 & 14 \\ \midrule
         Observed & $\{$All 14$\}$ & $\{$Age, & $\{$Age, & $\{$Age$\}$ & $\emptyset$ \\
                  &         & Neuro. diag., & Neuro. diag. \\
                  &         & Trauma diag., & Surgery$\}$  \\
                  &         & Surgery$\}$ &  \\
         \bottomrule
    \end{tabularx}}
    \caption{\textbf{Data settings considered during expert clinician evaluation.} Physicians are asked to rank action choices based on only state information ($\Gamma \approx 200$), or with varying amounts of observed contextual information ($\{$All 14$\}$ refers to all possible $\Zspace$ variables outlined in \Cref{tab:hirid_details}). }
    \label{tab:expert_eval_gamma}
\end{table}

We contacted our local institution's ethics committee to enquire about the possible necessity of ethics approval for this experimental framework. We were informed that this was not considered necessary as the experts contribute to the validation of algorithms and are thus not themselves the subject of the research, and as the undertaking comes with minimal risks to those experts (anonymous data collection). Best practice was nonetheless observed, by providing participants with an information and consent letter to inform them of their rights and obligations, and of how their data is collected and used. Participants were asked to read and sign this letter before collecting their anonymous expert opinion.

Results in \Cref{fig:hirid_res_expert} report the preference of clinicians for actions from either Delphic ORL or CQL or from behaviour cloning. We note their overall preference for the Delphic ORL policy in the confounded settings (high $\Gamma$). As more contextual information about the patient becomes available, however, and confounding is less marked (small $\Gamma$), physicians favour the behaviour cloning policy -- closer to expected clinical practice.

\section{Ablations and Additional Results} \label{appendix:extra_results}

\subsection{Sepsis Environment}

\begin{figure}[htb]
    \centering
    \begin{subfigure}[b]{0.43\textwidth}
        \centering
        \includegraphics[width=\textwidth]{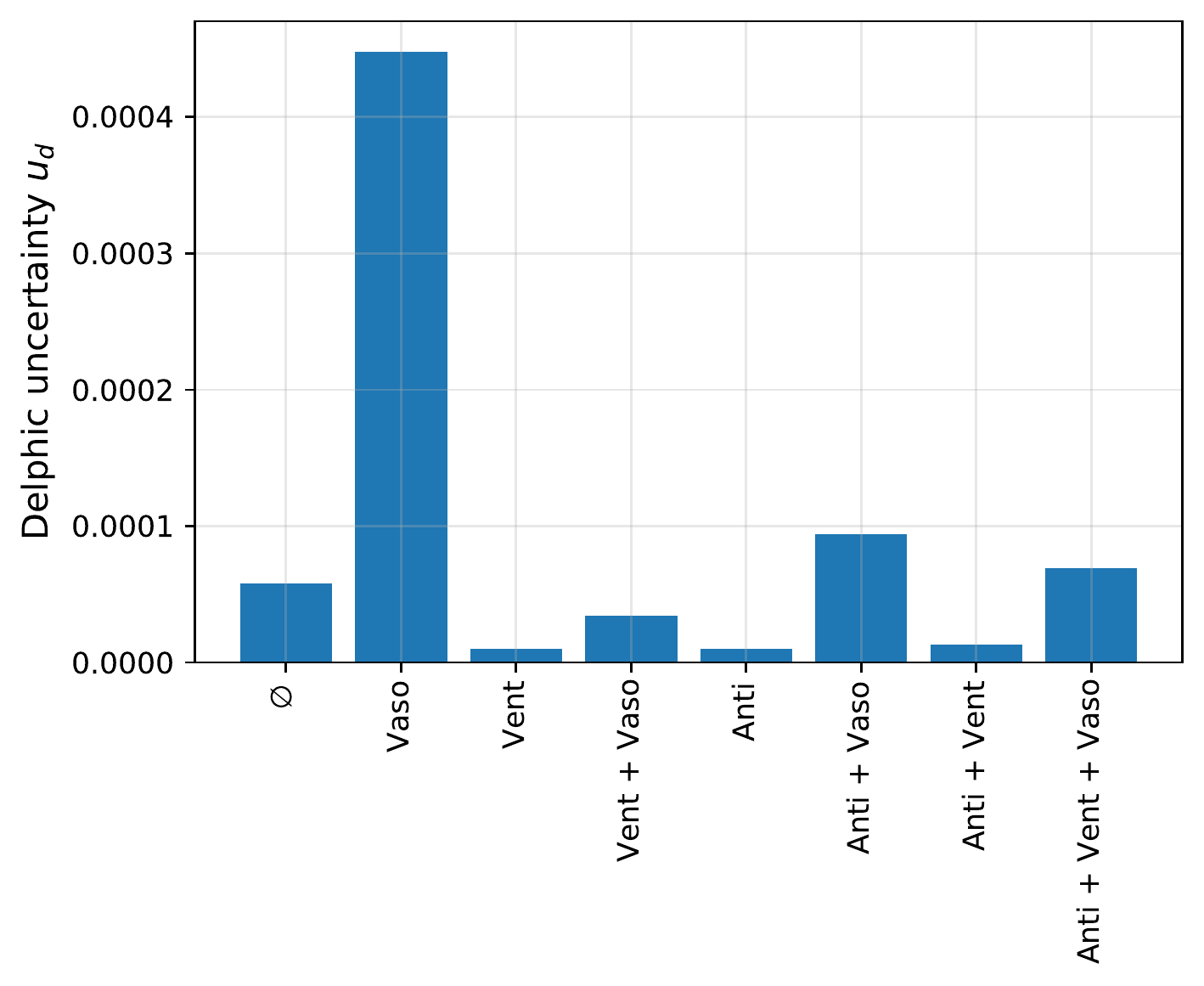}
        \caption{$u_d$ as a function of $a$.} \label{fig:u_vs_sa_sepsis}
    \end{subfigure} \hfill
    \begin{subfigure}[b]{0.43\textwidth}
        \centering
        \raisebox{3em}{\includegraphics[width=\textwidth]{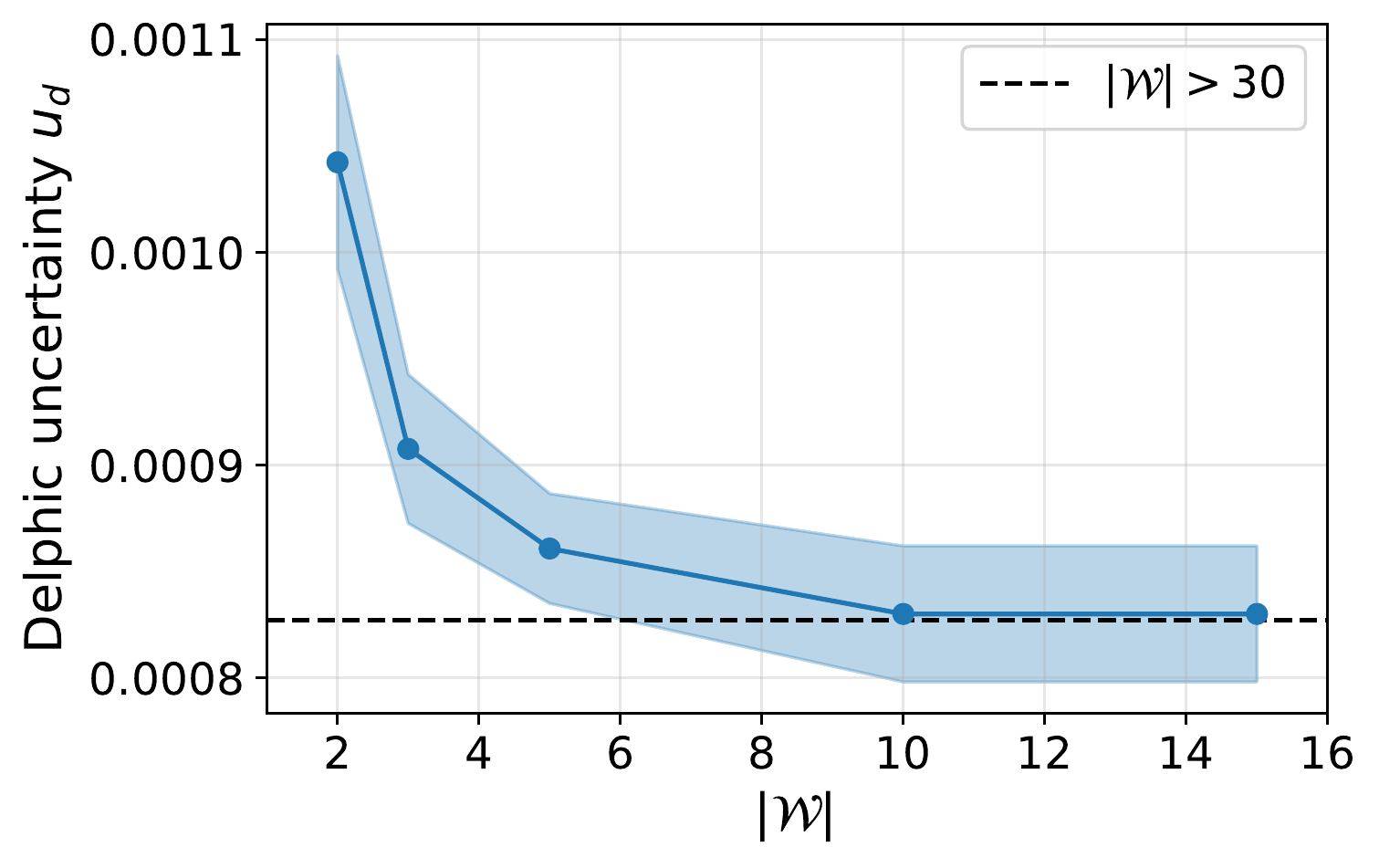}}
        \caption{$u_d^{\pi}$ as a function of $|\W|$.} \label{fig:u_vs_W_sepsis}
    \end{subfigure}     
    \caption{\textbf{Ablation Study: Delphic Uncertainty.} (a) Delphic uncertainty is highest under vasopressors in the sepsis environment, correctly identifying their confounded effect (Abbreviations: Vaso = Vasopressors, Anti = Antibiotics, Vent = Ventilation). (b) Empirically, only a small number of compatible worlds (for sepsis, $|\W| \approx 10$) is necessary to obtain an asymptotic estimate of $u_d$.}
    \label{fig:ablation_sepsis}
\end{figure}

\paragraph{Ablation Study: Delphic Uncertainty.} In \Cref{fig:u_vs_sa_sepsis}, we find that delphic uncertainty is highest on the sepsis dataset when treatment involves vasopressors. By design of the simulation \citep{Oberst2019}, this treatment is the only one for which patient evolution is confounded by the hidden diabetic status, which further supports the conclusion that delphic uncertainty captures model bias due to hidden confounding. In \Cref{fig:u_vs_W_sepsis}, we note that a only small number of world models (for sepsis, $|\W| \approx 10$) is necessary to obtain an estimate of delphic uncertainty consistent with a large number of world models. This motivates our practical choice to only consider a small set of world models to obtain a reasonable estimate of uncertainty for Delphic ORL, but warrants further theoretical work establishing guarantees and probability of correctness.

\begin{figure}[htb]
    \centering
    \begin{floatrow}
    \ffigbox{%
        \centering
        \includegraphics[width=0.4\textwidth]{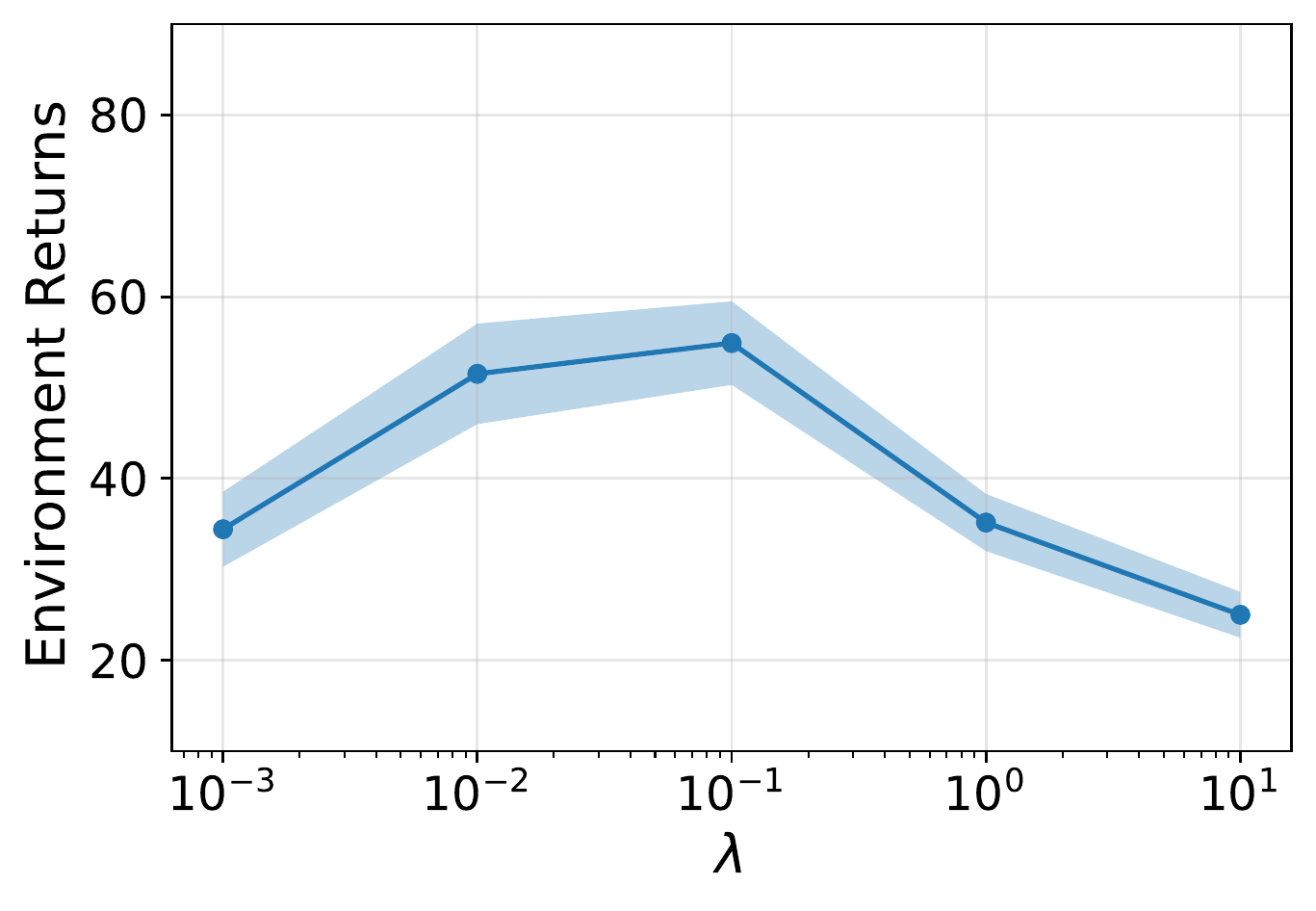}%
    }{%
        \caption{\textbf{Performance results as a function of hyperparameter $\lambda$} on the sepsis environment ($\Gamma=46$).} \label{fig:returns_vs_lambda_sepsis}
    }
    \capbtabbox{%
        \resizebox{0.45\textwidth}{!}{
        \begin{tabular}{lc}
        \toprule
        Algorithm & Environment Returns \\
        \midrule
        Online RL & 67.8 $\pm$ 1.1 \\
        \midrule
        BC & 38.5 $\pm$ 4.5 \\
        BCQ & 18.5 $\pm$ 2.4 \\
        CQL & 31.1 $\pm$ 3.5\\
        \midrule
        Delphic ORL ($u_d$ Threshold) & 24.6 $\pm$ 3.4\\
        Delphic BC (Weighting) &39.6 $\pm$ 4.1\\
        Delphic ORL (Weighting) & 44.7 $\pm$ 4.2\\
        Delphic ORL (Algo. \ref{alg:algo_qtarget}) & \textbf{54.9} $\pm$ 4.6\\
        \bottomrule
        \end{tabular}}
    }{%
          \caption{\textbf{Performance of different pessimism methods} on the sepsis environment ($\Gamma = 46$).} \label{tab:perf_results}%
    }
\end{floatrow}
\end{figure}

\paragraph{Ablation Study: Delphic ORL.} In \Cref{fig:returns_vs_lambda_sepsis}, we study the performance of Delphic ORL as a function of hyperparameter $\gamma$, interpolating between a naive implementation of Offline RL for very low values of $\gamma$ (virtually no penalty) and an excessively pessimistic algorithm, where the confounding penalty overcomes any possible high-reward behaviour. 

Next, \Cref{tab:perf_results} compares the performance of different approaches to implement pessimism with respect to delphic uncertainty. We find that our approach proposed in the main paper, based on penalising the target for the Bellman update, performs best in this experimental setting (sepsis dataset with $\Gamma = 46$). Weighting-based approaches also show promising performance (either matching or improving the performance of BC and CQL, respectively), which may be an avenue for further work and fine-tuning. Modifying the Bellman target to only include actions below a certain uncertainty threshold was however found to be excessively pessimistic, and degraded performance compared to the base CQL algorithm. Model-based Offline RL and Delphic ORL were not included as their performance was never found to improve over a random baseline policy. We hope this ablation study will motivate further work into the best possible approach to implement pessimism with respect to delphic uncertainty, to learn offline RL policies that are robust to hidden confounding bias.

\subsection{Real-World Clinical Dataset}

In this section, we provide additional evaluation metrics and investigations to understand the treatment strategies identified by the different algorithms considered, and in particular how Delphic ORL determines confounding-robust policies.

\begin{table}[h]
\caption{\textbf{Difference in action choices from $\dataset_{test}$}  across different algorithms (\%). Our method learns a distinct policy from the doctors'. Mean and 95\% CIs over 10 runs. Highest and overlapping values in bold.} \label{tab:difference_action_hirid}
    \resizebox{!}{3cm}{
    \begin{tabular}{lcccc}
\toprule
Confounders $\Zspace$   & BCQ            & BC             & CQL                     & \textbf{Delphic ORL}    \\ \midrule
All below               & \textbf{32.2} $\pm$ 1.3 & 19.7 $\pm$ 1.1 & \textbf{33.4} $\pm$ 0.9 & \textbf{35.2} $\pm$ 1.5 \\
$\{$age$\}$             & \textbf{31.5} $\pm$ 1.3 & 12.8 $\pm$ 0.4 & 27.3 $\pm$ 0.3          & \textbf{32.3} $\pm$ 0.5 \\
$\{$neuro. diag.$\}$    & 31.1 $\pm$ 1.3 & 16.3 $\pm$ 1.0 & \textbf{34.3} $\pm$ 1.3 & 30.6 $\pm$ 1.1          \\
$\{$gastro. diag.$\}$   & \textbf{27.1} $\pm$ 1.1 & 14.3 $\pm$ 0.9 & \textbf{28.9} $\pm$ 1.1          & \textbf{29.4} $\pm$ 1.3          \\
$\{$trauma$\}$          & \textbf{30.1} $\pm$ 1.5 & 12.8 $\pm$ 0.4 & \textbf{24.2} $\pm$ 0.4 & 22.2 $\pm$ 0.7          \\
$\{$cardio. diag.$\}$   & {28.7} $\pm$ 1.3 & 18.8 $\pm$ 1.2 & \textbf{36.2} $\pm$ 1.3 & {29.6} $\pm$ 1.6          \\
$\{$endo. diag.$\}$     & \textbf{27.4} $\pm$ 1.3 & 13.5 $\pm$ 0.8 & \textbf{27.3} $\pm$ 0.9 & 23.1 $\pm$ 0.9          \\
$\{$hemato. diag.$\}$   & \textbf{30.1} $\pm$ 1.5 & 12.4 $\pm$ 0.8 & {24.4} $\pm$ 1.1 & {23.6} $\pm$ 0.8 \\
$\{$weight$\}$          & \textbf{28.9} $\pm$ 1.3 & 13.2 $\pm$ 0.4 & {25.4} $\pm$ 0.6 & {23.6} $\pm$ 1.2 \\
$\{$sedation$\}$        & \textbf{30.5} $\pm$ 1.5 & 14.5 $\pm$ 0.7 & 25.1 $\pm$ 1.1          & 25.8 $\pm$ 1.0          \\
$\{$resp. diag.$\}$     & \textbf{27.7} $\pm$ 1.3 & 14.2 $\pm$ 0.6 & 28.5 $\pm$ 1.1          & \textbf{25.2} $\pm$ 1.2          \\
$\{$intoxication$\}$    & \textbf{25.7} $\pm$ 1.1 & 12.6 $\pm$ 0.6 & 26.3 $\pm$ 0.6          & 23.1 $\pm$ 0.9          \\
$\{$surgical status$\}$ & \textbf{27.3} $\pm$ 1.3 & 14.3 $\pm$ 0.6 & 23.9 $\pm$ 0.8          & 22.1 $\pm$ 1.2          \\
$\{$ortho. diag.$\}$    & \textbf{25.6} $\pm$ 1.1 & 12.3 $\pm$ 0.6 & 24.1 $\pm$ 1.1          & 22.3 $\pm$ 1.0          \\
$\{$sepsis$\}$           & \textbf{26.1} $\pm$ 1.1 & 15.6 $\pm$ 0.8 & 23.5 $\pm$ 0.8          & 21.9 $\pm$ 1.2          \\
\midrule
$\emptyset$             & \textbf{25.3} $\pm$ 0.9 & 12.2 $\pm$ 0.4 & 23.1 $\pm$ 0.8          & 21.7 $\pm$ 0.9        \\
\bottomrule
\end{tabular}
}
\end{table}

\Cref{tab:difference_action_hirid} provides a quantitative analysis of the disparities in action choices between different algorithms and the doctors' policy. As expected, behaviour cloning exhibits the closest resemblance to the doctors' treatment policy, which aligns with the characteristics of observational datasets. However, our proposed method outperforms behaviour cloning in terms of learning a distinct policy that deviates from the doctors' actions. These findings highlight the unique capabilities of our method in capturing important features and patterns beyond the direct imitation of doctors, enabling the model to make informed decisions that may differ from the observational data and potentially lead to improved treatment outcomes.

Following published recommendations on evaluating RL models in observational settings \citep{gottesman2018evaluating}, we also analyse where policies differ most from the action choices in the observational dataset, and find that the policy learned by Delphic ORL diverges most at high disease severity (SOFA scores $\approx$ 15-20). In these cases, our policy appears to prescribe less fluids and vasopressors than in the data -- which may be reasonable if unsure about possible adverse effects of an intervention. This relates to a comment received from one of the expert clinicians interviewed: ``If I lack information about a patient [e.g. age, medical background and deliberately excluded variables], I would probably be more conservative with my treatment''. Finally, we note a closer match to actions in the observational data at very high disease severity (SOFA score $>20$), where negative rewards for \textit{not} taking a therapeutic action outweighs potential confounding bias. Beyond this analysis, further insights could be gained by comparing interpretable representations of the trained policies \citep{pace2022}, but we leave this as further work.

\end{document}